\documentclass[letterpaper]{article} 
\usepackage{aaai2026}  
\usepackage{times}  
\usepackage{helvet}  
\usepackage{courier}  
\usepackage[hyphens]{url}  
\usepackage{graphicx} 
\usepackage{natbib}  
\usepackage{caption} 
\usepackage{algorithm}
\usepackage{algorithmic}

\usepackage{newfloat}
\usepackage{listings}
\usepackage{booktabs}

\usepackage{graphicx}
\usepackage{amsmath}
\usepackage{amssymb}
\usepackage{booktabs}

\usepackage{subcaption} 

\DeclareCaptionStyle{ruled}{labelfont=normalfont,labelsep=colon,strut=off} 
\floatstyle{ruled}
\newfloat{listing}{tb}{lst}{}
\floatname{listing}{Listing}
\title{Phys-Liquid: A Physics-Informed Dataset for Estimating 3D Geometry and Volume of Transparent Deformable Liquids}

\author{
    Ke Ma\textsuperscript{\rm 1},
    Yizhou Fang\textsuperscript{\rm 2},
    Jean\mbox{-}Baptiste Weibel\textsuperscript{\rm 3},
    Shuai Tan\textsuperscript{\rm 4},
    Xinggang Wang\textsuperscript{\rm 5},
    Yang Xiao\textsuperscript{\rm 6},
    Yi Fang\textsuperscript{\rm 7,8},
    Tian Xia\textsuperscript{\rm 2}\thanks{Corresponding author.}
}
\affiliations{
    \textsuperscript{\rm 1}School of Artificial Intelligence and Automation, Huazhong University of Science and Technology\\
    \textsuperscript{\rm 2}School of Software and Engineering, Huazhong University of Science and Technology\\
    \textsuperscript{\rm 3}Human-Centered AI Lab, Institute of Forest Engineering, BOKU University\\
    \textsuperscript{\rm 4}Shanghai Jiao Tong University\\
    \textsuperscript{\rm 5}School of Electronic Information and Communications, Huazhong University of Science and Technology\\
    \textsuperscript{\rm 6}National Key Laboratory of Multispectral Information Intelligent Processing Technology,\\
    School of Artificial Intelligence and Automation, Huazhong University of Science and Technology\\
    \textsuperscript{\rm 7}Center for Artificial Intelligence and Robotics, New York University Abu Dhabi\\
    \textsuperscript{\rm 8}Embodied AI and Robotics (AIR) Lab, New York University\\
    \{make, fangyizhou, xgwang, Yang\_Xiao, tianxia\}@hust.edu.cn, 
    jb.weibel@gmail.com, tanshuai0219@stju.edu.cn, yfang@nyu.edu
}
\usepackage{bibentry}

\begin{document}

\maketitle

\begin{abstract}
Estimating the geometric and volumetric properties of transparent deformable liquids is challenging due to optical complexities and dynamic surface deformations induced by container movements. Autonomous robots performing precise liquid manipulation tasks—such as dispensing, aspiration, and mixing—must handle containers in ways that inevitably induce these deformations, complicating accurate liquid state assessment. Current datasets lack comprehensive physics-informed simulation data representing realistic liquid behaviors under diverse dynamic scenarios. To bridge this gap, we introduce Phys-Liquid, a physics-informed dataset comprising 97,200 simulation images and corresponding 3D meshes, capturing liquid dynamics across multiple laboratory scenes, lighting conditions, liquid colors, and container rotations. To validate the realism and effectiveness of Phys-Liquid, we propose a four-stage reconstruction and estimation pipeline involving liquid segmentation, multi-view mask generation, 3D mesh reconstruction, and real-world scaling. Experimental results demonstrate improved accuracy and consistency in reconstructing liquid geometry and volume, outperforming existing benchmarks. The dataset and associated validation methods facilitate future advancements in transparent liquid perception tasks. The dataset and code are available at https://dualtransparency.github.io/Phys-Liquid/.
\end{abstract}

\begin{figure}[h]
    \centering
    \includegraphics[width=0.47\textwidth]{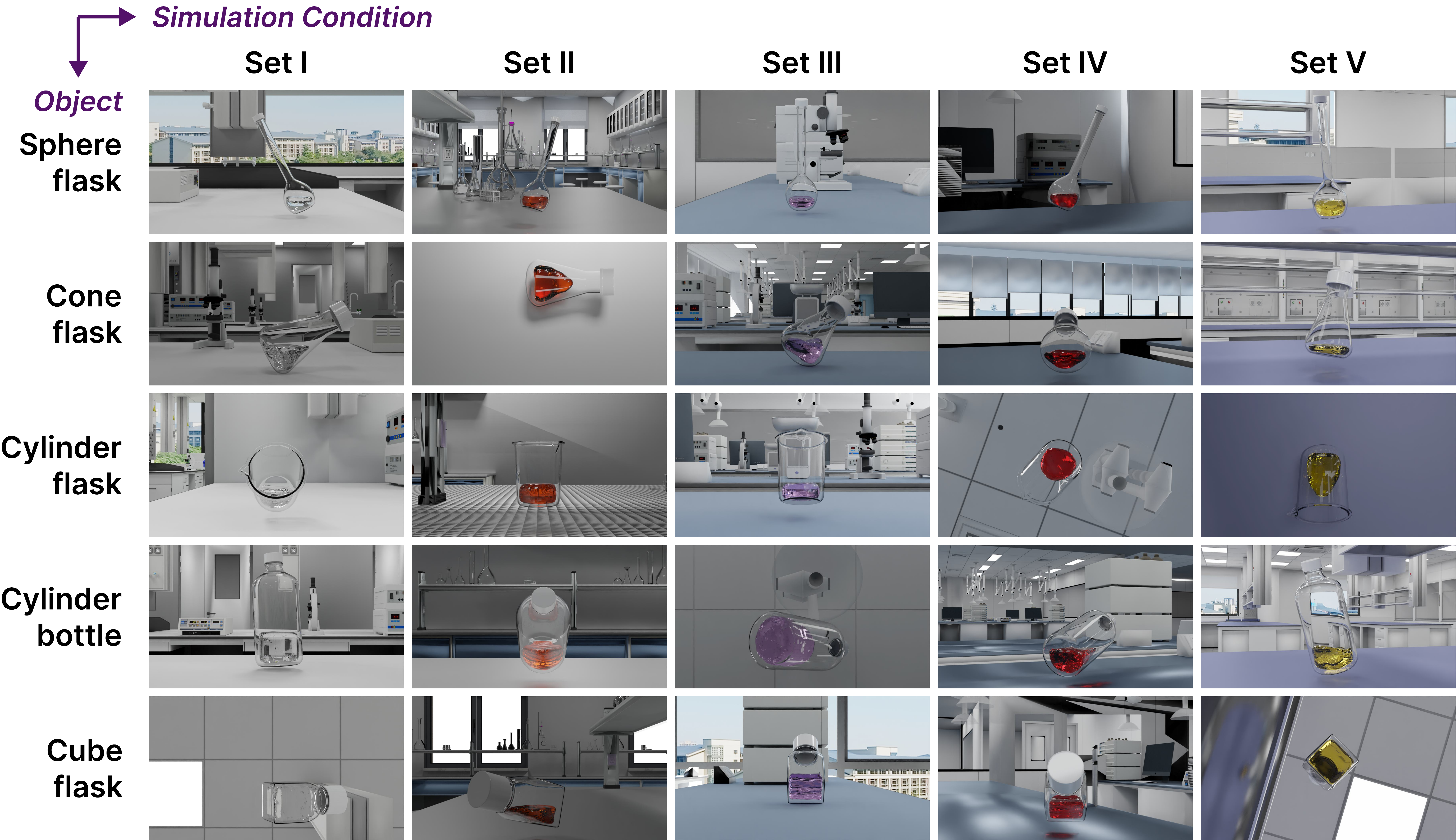}
    \caption{Simulation samples of five transparent containers under five distinct simulation sets from Phys-Liquid, showing variations in laboratory scenes, lighting conditions, container rotations, liquid colors, volumes, and deformations.}
    \label{fig:fig1}
\end{figure}

\section{Introduction}
\label{sec:intro}

Understanding the geometric and volumetric properties of transparent deformable liquids is essential for embodied robots~\cite{liu2024aligning} operating in autonomous laboratory environments. As robotics and large multimodal models~\cite{radford2021learning, li2023blip, driess2023palm} advance, autonomous robots are increasingly tasked with handling complex procedures in biomedical and biochemical experiments~\cite{dai2024autonomous, szymanski2023autonomous, triantafyllidis2023hybrid, xie2023inverse, boiko2023autonomous, burger2020mobile}. Precise liquid manipulation tasks-such as dispensing, aspiration, and mixing-require these systems to accurately perceive dynamic liquid deformation caused inevitably by container movements. Without robust and realistic simulation datasets that represent the complexities of deformable liquids, accurately assessing these liquids remains a significant challenge, potentially causing experimental failures and resource losses.

Current datasets are insufficient for addressing the complex challenges posed by transparent deformable liquids in dynamic laboratory environments. Existing large-scale 3D datasets like Objaverse~\cite{deitke2023objaverse} primarily include rigid and opaque objects, with limited representation of liquid-filled transparent containers. Datasets such as ClearGrasp~\cite{sajjan2020clear}, ClearPose~\cite{chen2022clearpose}, and TODD \cite{fang2022transcg} focus on transparent object perception but neglect the contained liquids. Although datasets like DTLD~\cite{wang2024towards} and the dataset by Narasimhan et al.~\cite{narasimhan2022self} include transparent liquids, they either provide static liquid states or limited views without simulating realistic deformation dynamics induced by container rotations. This lack of dynamic realism restricts the development of robust algorithms for precise robotic manipulation in real-world scenarios.

To address these limitations, we introduce Phys-Liquid, a physics-informed simulation dataset specifically designed to capture the dynamic behaviors of transparent deformable liquids under realistic laboratory conditions. Unlike previous datasets, Phys-Liquid systematically simulates liquid deformations induced by container rotations based on precise physical modeling governed by Navier-Stokes equations \cite{chorin1968numerical}. We proposed a dataset generation workflow using Blender~\cite{blender2018blender} that includes defining diverse laboratory scenes, lighting conditions, liquid properties, and systematically rotating containers to induce temporal liquid deformations, subsequently rendered from multiple orthographic viewpoints. By comparing simulated liquid behaviors against corresponding real-world captures, we further validate the realism and fidelity of our simulation data. Our dataset comprises 97,200 simulation images generated from 20 common laboratory containers, encompassing diverse laboratory scenes, liquid colors, lighting conditions, and rotation modes. The dataset includes annotated 3D liquid meshes, providing precise geometric and volumetric information for rigorous validation and benchmarking.

A key aspect of validating the realism and effectiveness of a physics-informed dataset is to demonstrate its utility in enabling accurate predictions of liquid geometry and volume from visual inputs. To this end, we developed a four-stage pipeline that leverages Phys-Liquid for reconstructing and estimating liquid geometry from single images. This pipeline includes segmentation of liquid regions, generation of multi-view liquid masks via diffusion models \cite{wang2025crm}, 3D mesh reconstruction using triplane methods \cite{wang2025crm}, and scaling reconstructed meshes to match real-world dimensions. By comparing the reconstructed liquid states against ground truth simulation meshes, we quantitatively assess the dataset’s realism and applicability to real-world manipulation tasks.

Experimental validation demonstrates that Phys-Liquid significantly improves the accuracy and consistency of liquid geometry reconstruction compared to existing benchmarks. Moreover, evaluation on real-world datasets confirms that models trained on Phys-Liquid exhibit strong generalization capabilities. These results underscore the dataset’s realistic representation of dynamic liquid behaviors and highlight its potential to substantially enhance future robotic manipulation and liquid perception research.

Our contributions are summarized as follows:
\begin{itemize}
\item We introduce Phys-Liquid, a physics-informed dataset addressing gaps in current liquid datasets by capturing realistic dynamic deformations, providing a foundational resource for future research in transparent liquid perception and precise liquid handling applications.
\item We develop a four-stage reconstruction pipeline to validate the dataset's effectiveness, demonstrating superior performance over existing benchmarks on both simulation and real-world liquid datasets.
\end{itemize}


\section{Related Work}
\label{sec:relatedwork}

\subsection{Liquid Datasets in Laboratory Scenes}

Existing large-scale 3D datasets like Objaverse~\cite{deitke2023objaverse} include various chemistry lab glassware (e.g., beakers, cylindrical flasks, measuring glasses, tubes, and conical flasks); however, most of these objects are typically empty and do not involve liquids. Transparent object datasets like TOD~\cite{liu2020keypose}, TODD~\cite{xu2021seeing}, TransCG~\cite{fang2022transcg}, ClearGrasp~\cite{sajjan2020clear}, StereOBJ-1M~\cite{liu2021stereobj}, and ClearPose~\cite{chen2022clearpose} mainly focus on 6D pose estimation for laboratory objects but generally do not represent liquids inside these containers. Gautham et al.~\cite{narasimhan2022self} introduced a dataset capturing transparent chemistry flasks with transparent liquids from a single camera viewpoint, lacking representation of dynamic liquid deformations. The DTLD dataset proposed by Wang et al.~\cite{wang2024towards} comprises 27,458 images depicting liquids in four biomedical flasks captured under multi-view laboratory settings; however, the dataset only includes static liquid states without deformation dynamics induced by container movements. Thus, these datasets do not sufficiently address the dynamic characteristics crucial for precise liquid perception in realistic robotic manipulation scenarios.

\subsection{Physics-Informed Liquid Simulation Datasets}

Physics-informed methods have been increasingly employed in simulating realistic deformable or fluid behaviors for various computer vision tasks~\cite{banerjee2024physics, lin2025phys4dgen, tang2024divide,yan2023bispd,li2025stitchfusion, wu2025enhancing}, leveraging fundamental physical principles to enhance dataset authenticity. Recent liquid simulation studies have been explored across diverse applications, including robotic manipulation~\cite{lin2023pourit, eppel2022predicting, moya2023thermodynamics, qian2024understanding}, river flow velocimetry evaluation~\cite{bodart2022synthetic}, liquid volume estimation~\cite{liu2023inferring}, surface detection~\cite{richter2022image}, and temporal prediction~\cite{wiewel2019latent}. Various liquid datasets have been generated using different simulation pipelines: for instance, TransProteus~\cite{eppel2022predicting} simulates liquid shapes inside transparent vessels using Blender~\cite{blender2018blender} Mantaflow module~\cite{thuerey2016mantaflow}. Richter et al.~\cite{richter2022image} simulated a fountain liquid dataset with Blender~\cite{blender2018blender} to reconstruct liquids from surface detections. However, existing physics-informed datasets generally do not simulate dynamic liquid deformations within transparent containers induced by physical manipulations, thus leaving a gap in realistic representation of laboratory liquid behaviors.

\subsection{Liquid Reconstruction Methods from Images}

Several methods have been proposed for reconstructing liquids from visual data to enable downstream estimation tasks. Eppel et al.~\cite{eppel2022predicting} reconstructed 3D liquid shapes from RGB images by predicting XYZ maps, though facing limitations in scenes lacking precise depth data for transparent liquids. Richter et al.~\cite{richter2022image} proposed reconstructing 3D liquid meshes from 2D surface detections by optimizing particle positions based on physical constraints to match detected liquid surfaces. Additionally, recent advancements in single-view 3D reconstruction~\cite{chan2022efficient, cheng2023sdfusion, gupta20233dgen, vahdat2022lion, zheng2023locally, liu2024make, wu2024unique3d, xu2024instantmesh, wang2025crm} have demonstrated high-quality mesh generation from single images, showing potential for liquid reconstruction tasks.

\section{Physics-Informed Dataset}
\label{sec:dataset}

We present Phys-Liquid, a physics-informed dataset capturing the dynamic deformation of transparent liquids within transparent containers under systematic rotational motions in laboratory scenes. Figure~\ref{fig:fig1} presents examples of five containers under comprehensive simulation conditions.

\subsection{Physics-Informed Liquid Simulation Method}

To simulate realistic liquid behaviors, we leverage the Navier-Stokes equations~\cite{chorin1968numerical}, which accurately describe the fundamental physics of fluid dynamics by accounting for velocity, pressure, viscosity, and external forces. Specifically, the fluid flow in our simulation is governed by the momentum equation:

    \begin{equation}
    \frac{D\mathbf{u}}{Dt} = -\frac{1}{\rho} \nabla p + \nu \nabla^2 \mathbf{u} + \mathbf{g},
    \label{eq:important}
    \end{equation}
    
    where $\mathbf{u}$ represents the fluid's velocity field, indicating the direction and speed of each particle, $\rho$ denotes fluid density, $p$ is the pressure field, dictating flow based on high and low-pressure regions, $\nu$ is the kinematic viscosity that models internal friction within the fluid, affecting how smoothly it flows and $\mathbf{g}$ represents external forces that impact fluid behavior. This equation describes the fluid acceleration $\frac{D\mathbf{u}}{Dt}$ as a result of internal pressure forces $\left(-\frac{1}{\rho} \nabla p\right)$, viscous forces $(\nu \nabla^2 \mathbf{u})$, and external forces $\mathbf{g}$.
    Additionally, we enforce the incompressibility condition to ensure constant fluid density and realistic liquid behavior:
    
    \begin{equation}
    \nabla \cdot \mathbf{u} = 0
        \label{eq:important}
    \end{equation}
    
    This constraint ensures that fluid density remains constant, maintaining the divergence-free nature of the velocity field. It enforces an accurate fluid behavior by preventing artificial compressions or expansions. We solve these equations using Mantaflow~\cite{thuerey2016mantaflow}, integrated within Blender~\cite{blender2018blender}, enabling accurate simulation of liquid dynamics under various container rotations.

\subsection{Data Generation Pipeline}

\begin{figure}[h]
    \centering
    \includegraphics[width=0.2\textwidth]{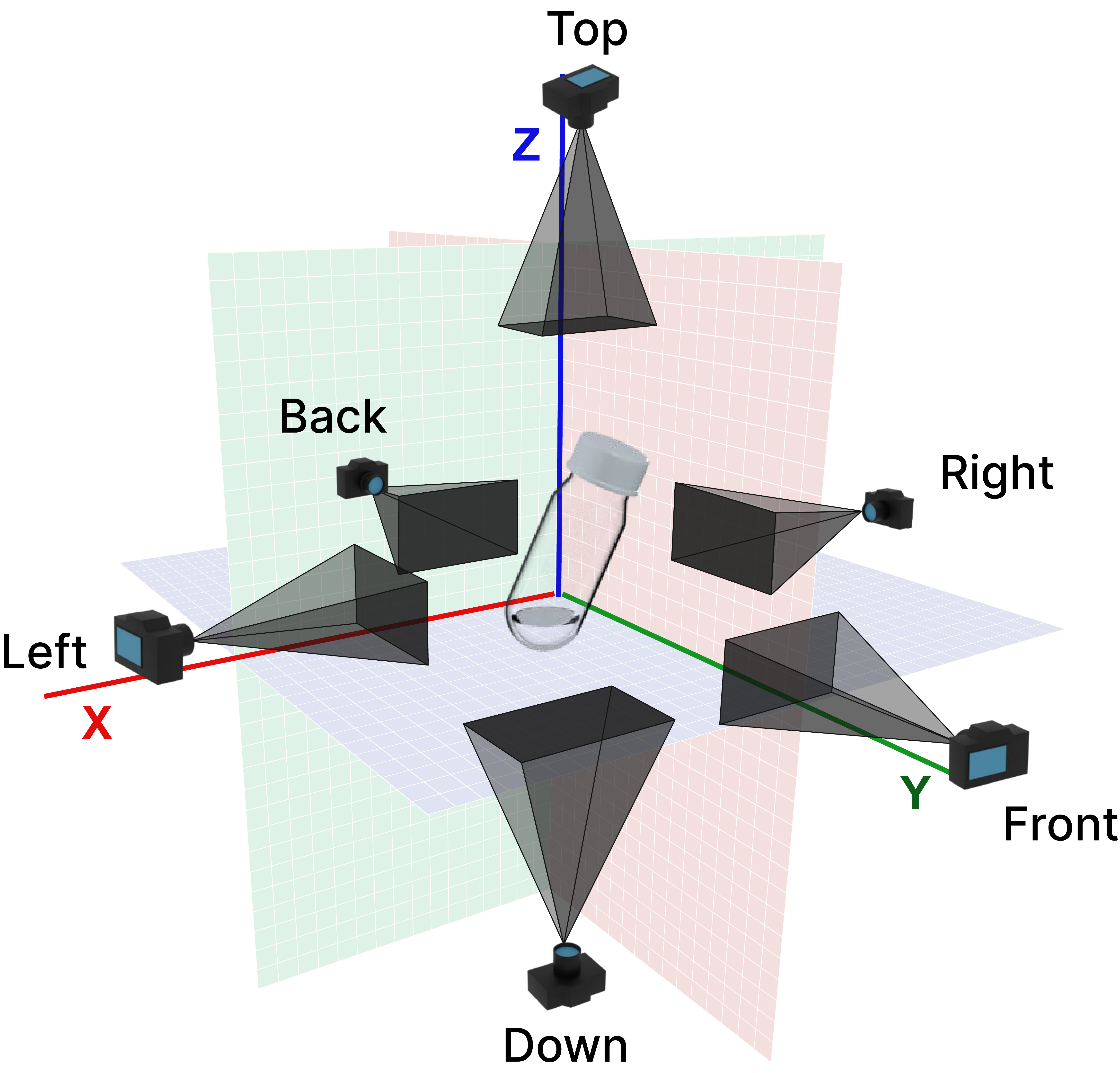}
    \caption{The setting of six orthographic camera views for the representation of triplane in the scene setting.}
    \label{fig:fig2}
\end{figure}

\paragraph{Scene Creation}
Our data generation workflow begins with constructing diverse laboratory scenes, including realistic lab environments, lighting variations, commonly-used laboratory containers, liquid colors, and controlled camera setups. Specifically, we design five realistic laboratory scenes containing experimental setups, lab equipment, and workbenches. We simulate eight indoor lighting conditions varying in intensity and orientation. We select twenty common transparent laboratory containers covering a variety of shapes (cubes, cylinders, cones, spheres, composite forms) and sizes, each with distinct materials and textures. Transparent liquids are simulated in five typical colors and various initial volumes. Six orthographic virtual cameras (top, bottom, front, back, left, right) are positioned to capture the liquid deformations and container rotations from multiple viewpoints, illustrated in Figure~\ref{fig:fig2}. Each simulation set randomly selects configurations, assigning five distinct conditions per container, resulting in 100 unique combinations.

\paragraph{Object Rotation Over Time}

Containers undergo rotational motions to induce liquid deformation. Rotation patterns are defined along the X, Y, and Z axes, creating six distinct combinations (excluding rotations solely around the Z-axis), with rotations ranging from 0\textdegree{} to 80\textdegree{}. We record 81 time frames for continuous motion, with each frame corresponding to a specific object pose and liquid deformation.

\paragraph{Liquid Simulation}

\begin{figure}[h]
    \centering
    \includegraphics[width=0.35\textwidth]{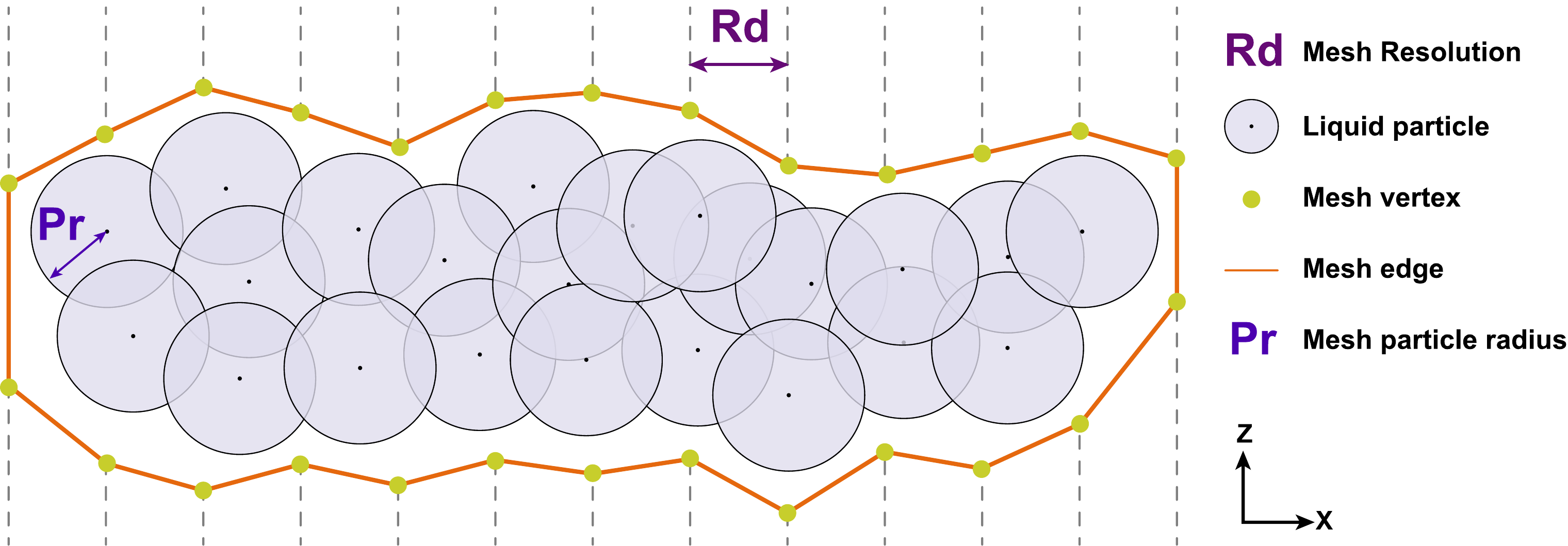}
    \caption{The liquid mesh formed by particles.}
    \label{fig:fig3}
\end{figure}

The liquid is represented by a particle system enclosed within a mesh defining the liquid's surface illustrated in Figure~\ref{fig:fig3}. Key parameters controlling the simulation include mesh resolution, particle radius, particle limits, and the FLIP~\cite{brackbill1988flip} ratio. As the container rotates, liquid particles interact with the inner surfaces, undergoing collisions governed by the Navier-Stokes equations~\cite{chorin1968numerical}. This results in realistic liquid deformations and varying surface meshes influenced by container geometry and motion.

\paragraph{Multi-View Rendering}

\begin{figure}[t]
    \centering
    \begin{subfigure}{0.48\textwidth}
        \centering
        \includegraphics[width=\textwidth]{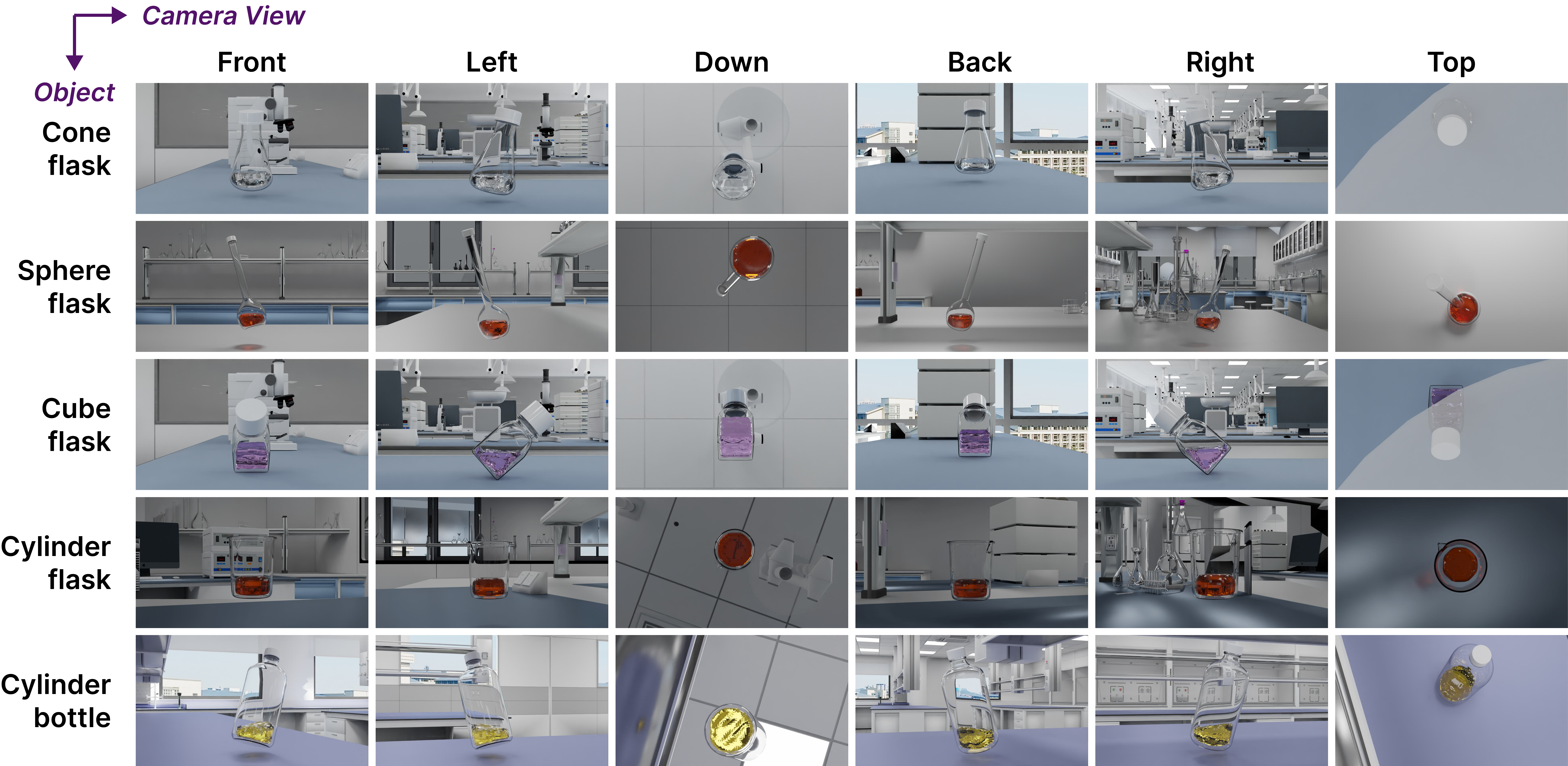}
        \caption{ Image samples of different objects captured simultaneously from six orthographic camera views at the same time frame.}
        \label{fig:fig4a}
    \end{subfigure}
    \hfill
    \begin{subfigure}{0.48\textwidth}
        \centering
        \includegraphics[width=\textwidth]{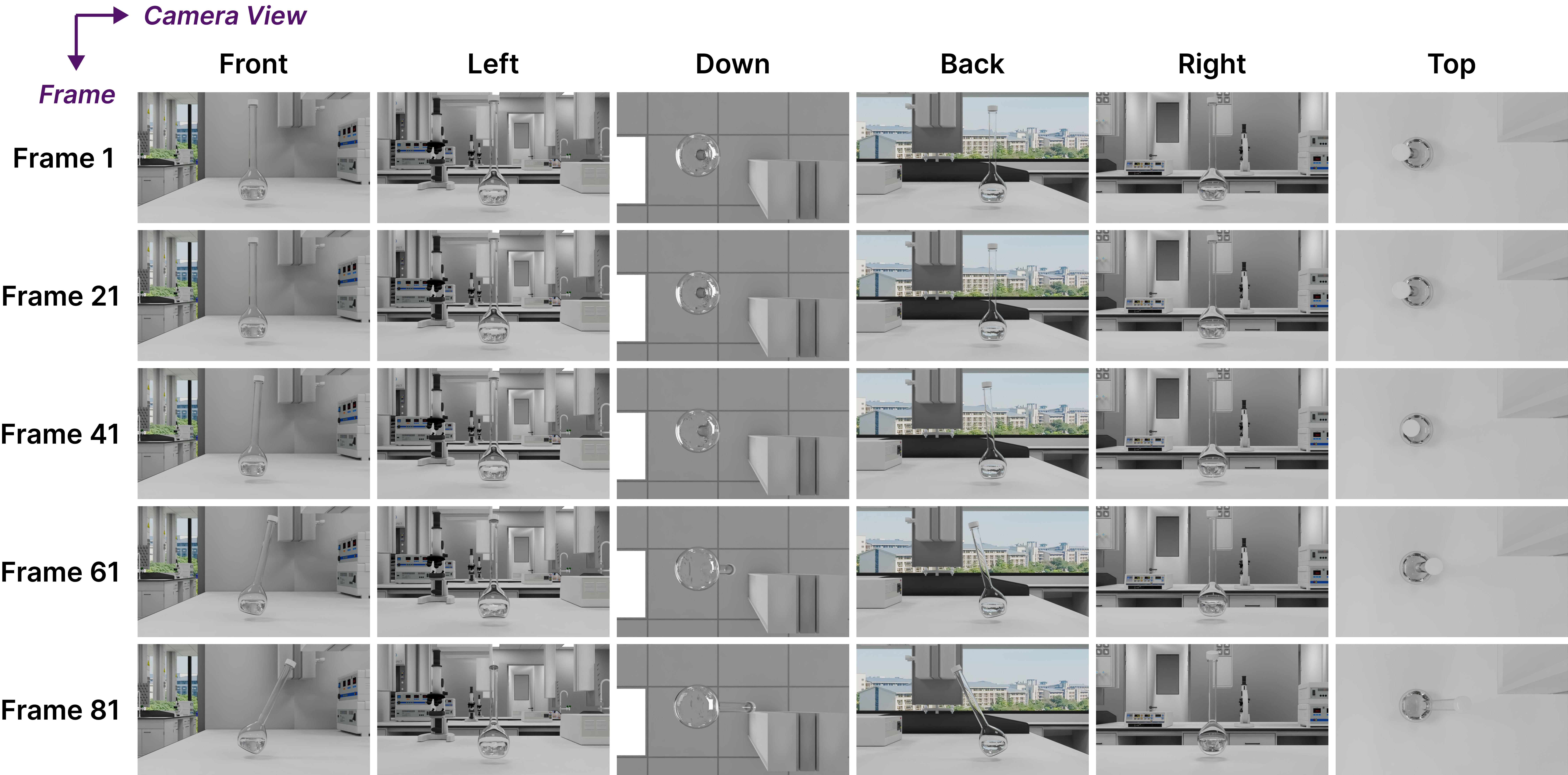}
        \caption{Image samples of the same object captured from six orthographic camera views across multiple time frames.}
        \label{fig:fig4b}
    \end{subfigure}
    
    \caption{Multi-view and temporal representations.}
    \label{fig:fig4}
\end{figure}

We render six orthographic views using virtual cameras to capture perspectives that are challenging to achieve in real-world settings. Images are rendered every frame over the sequence of 81 frames capturing continuous motion. For each time frame, we generate scene images (container, liquid, background), liquid-only masks, and liquid meshes saved in OBJ format with calibrated real-world dimensions. Figure~\ref{fig:fig4a} demonstrates multi-view images for different containers at one time frame, and Figure~\ref{fig:fig4b} shows temporal variation at 20-frame intervals across multiple views for a single container.

\subsection{Comparison with Real-World Captures}

\begin{figure}[h]
    \centering
    \includegraphics[width=0.45\textwidth]{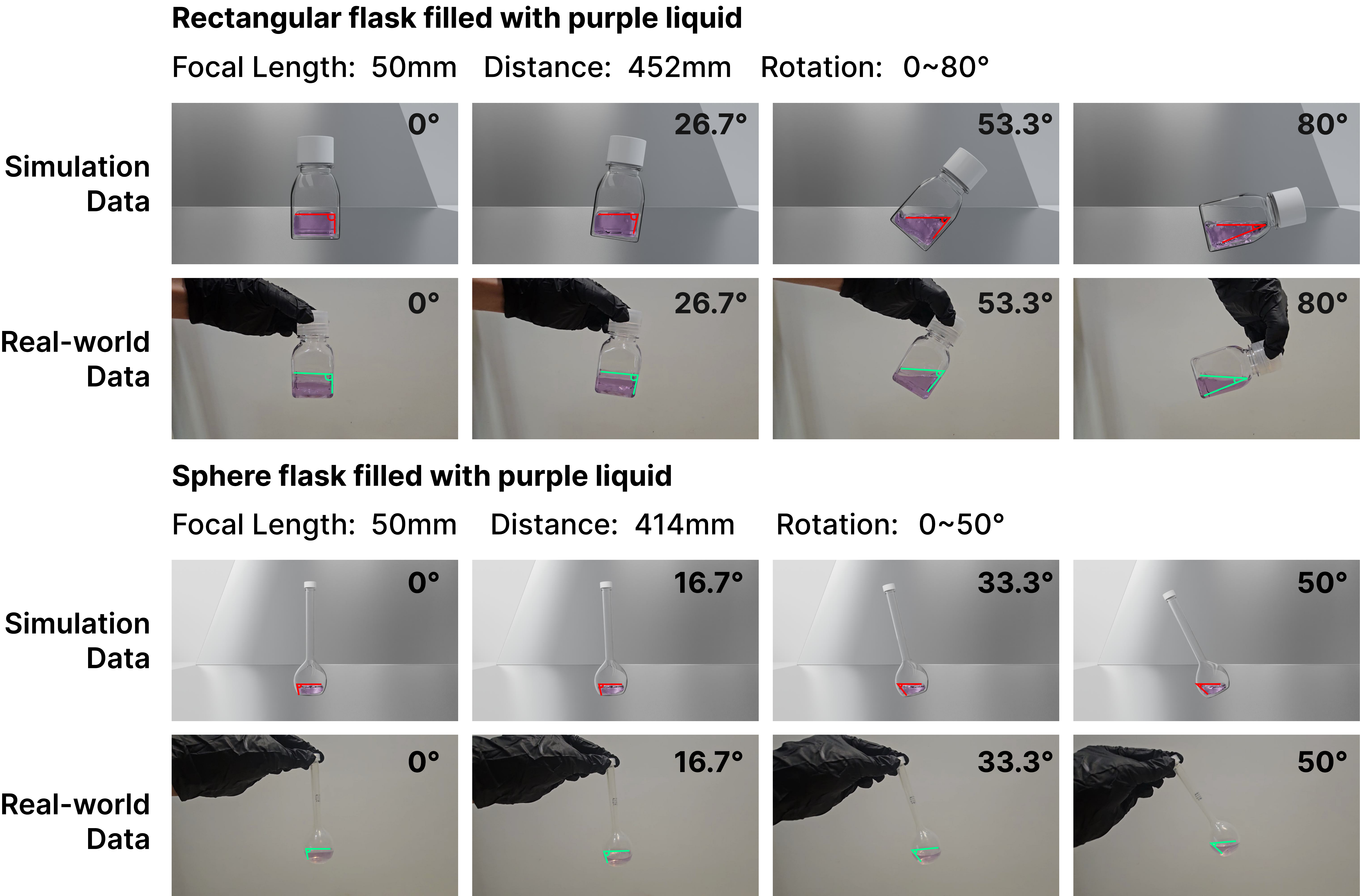}
    \caption{Validation of simulation realism by comparing liquid deformations with real-world experiments.}
    \label{fig:fig5}
\end{figure}

To further validate the dataset realism, we performed comparative experiments between simulated and real-world liquid behaviors. For ten representative containers from the Phys-Liquid dataset, we recorded liquid deformations in physical experiments under comparable rotational conditions. We analyzed visual and physical properties, such as liquid deformation, flow patterns, and interactions with container surfaces. In Figure~\ref{fig:fig5}, the angles between the liquid's top surface and container side walls are annotated in red for simulation data and green for real-world data regarding two objects. These highlighted angle pairs at each rotational pose closely align, demonstrating that our simulations approximate real-world liquid deformation behaviors.

\subsection{Dataset Analysis and Visualization}

Phys-Liquid comprises 97,200 simulation images (scene images and liquid masks) across 100 simulation configurations, 8,100 annotated OBJ-format liquid meshes with real-world size, and CAD models of 20 laboratory containers.

\begin{figure*}[t]
    \centering
  \begin{subfigure}{0.24\linewidth}
    \includegraphics[width=1\linewidth]{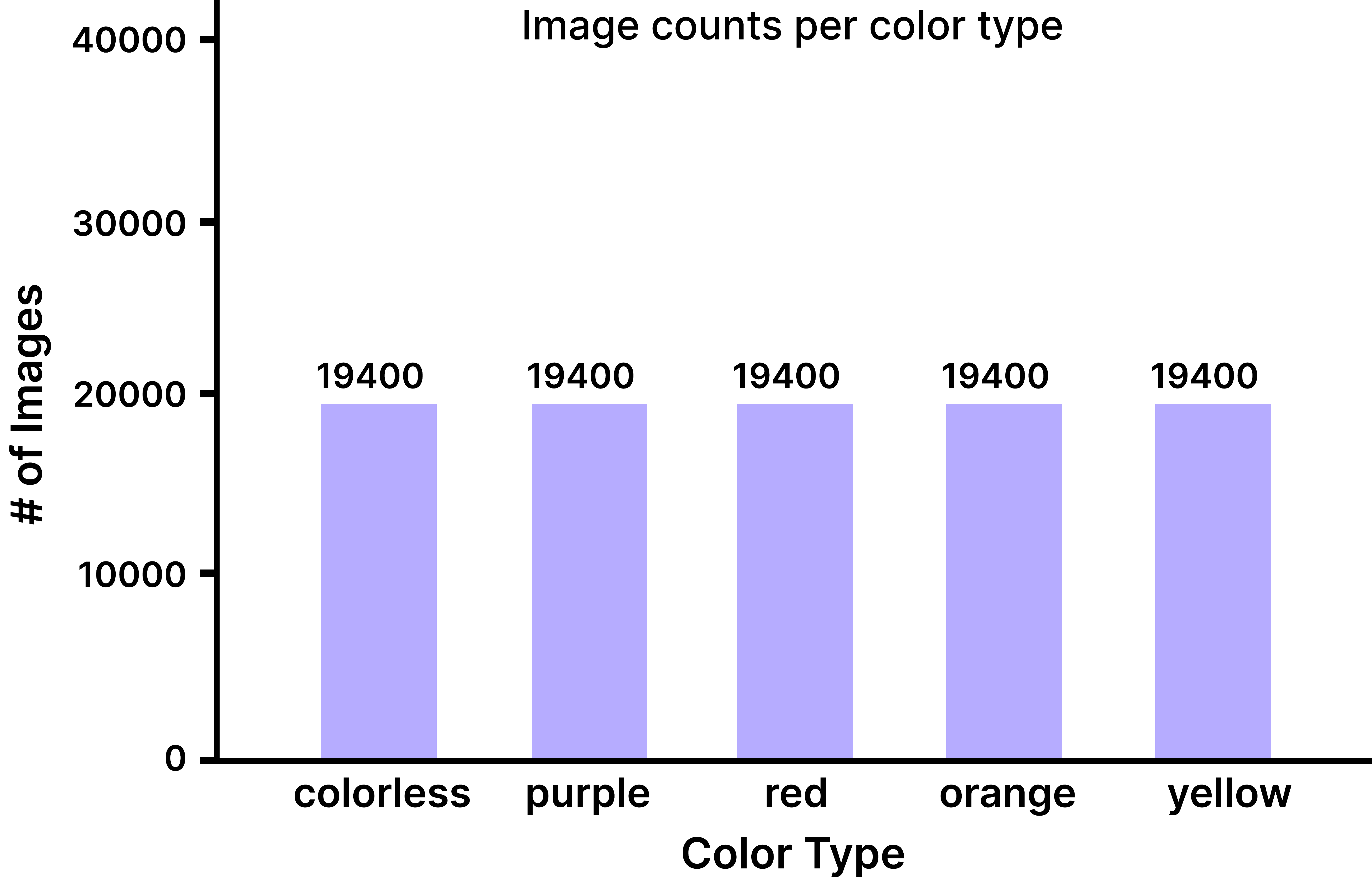}
    \caption{The number of images for each liquid color.}
    \label{fig:fig6a}
  \end{subfigure}
  \begin{subfigure}{0.24\linewidth}
    \includegraphics[width=1\linewidth]{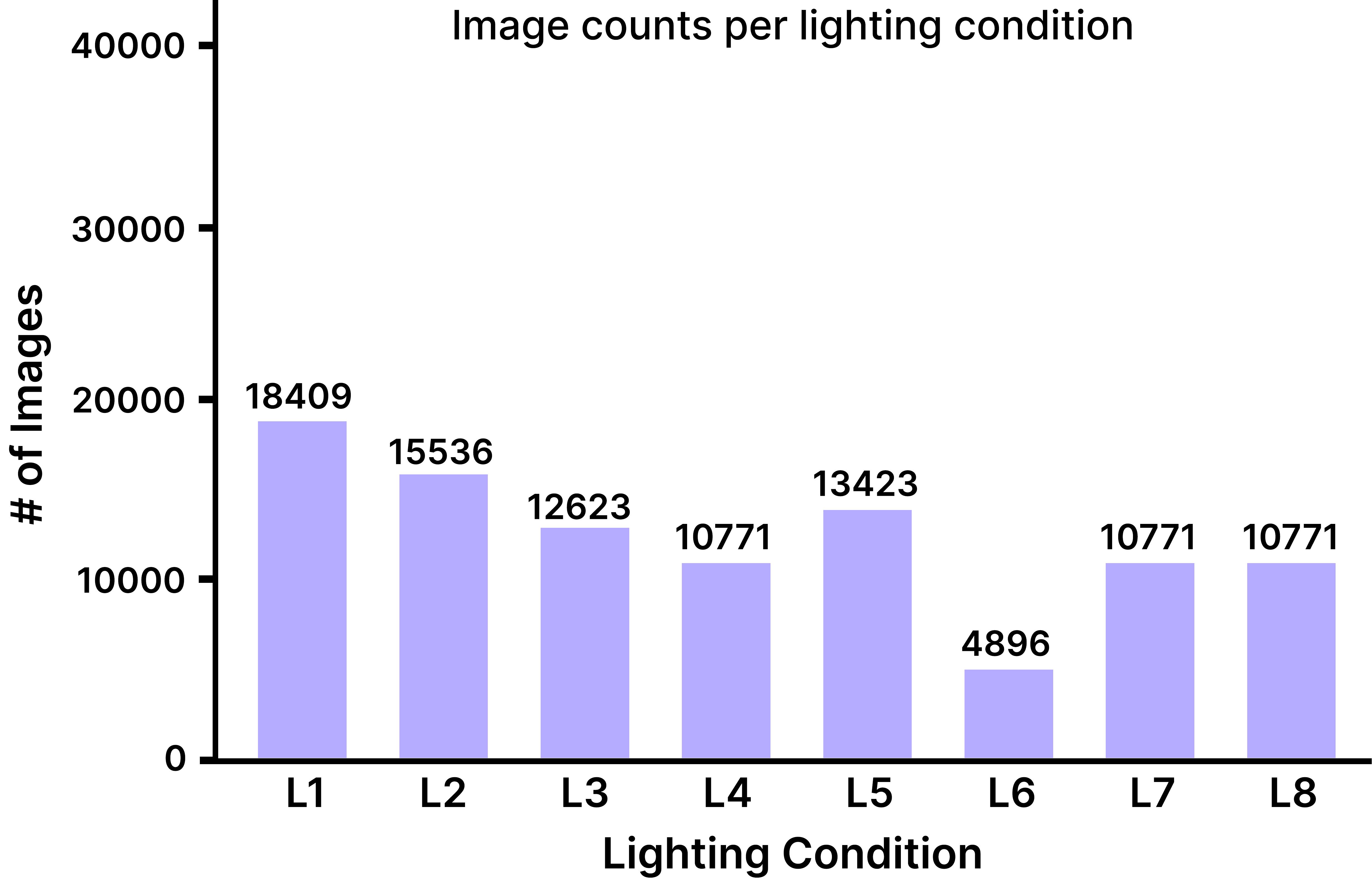}
    \caption{The number of images under each lighting condition.}
    \label{fig:fig6b}
  \end{subfigure}
  \begin{subfigure}{0.24\linewidth}
\includegraphics[width=1\linewidth]{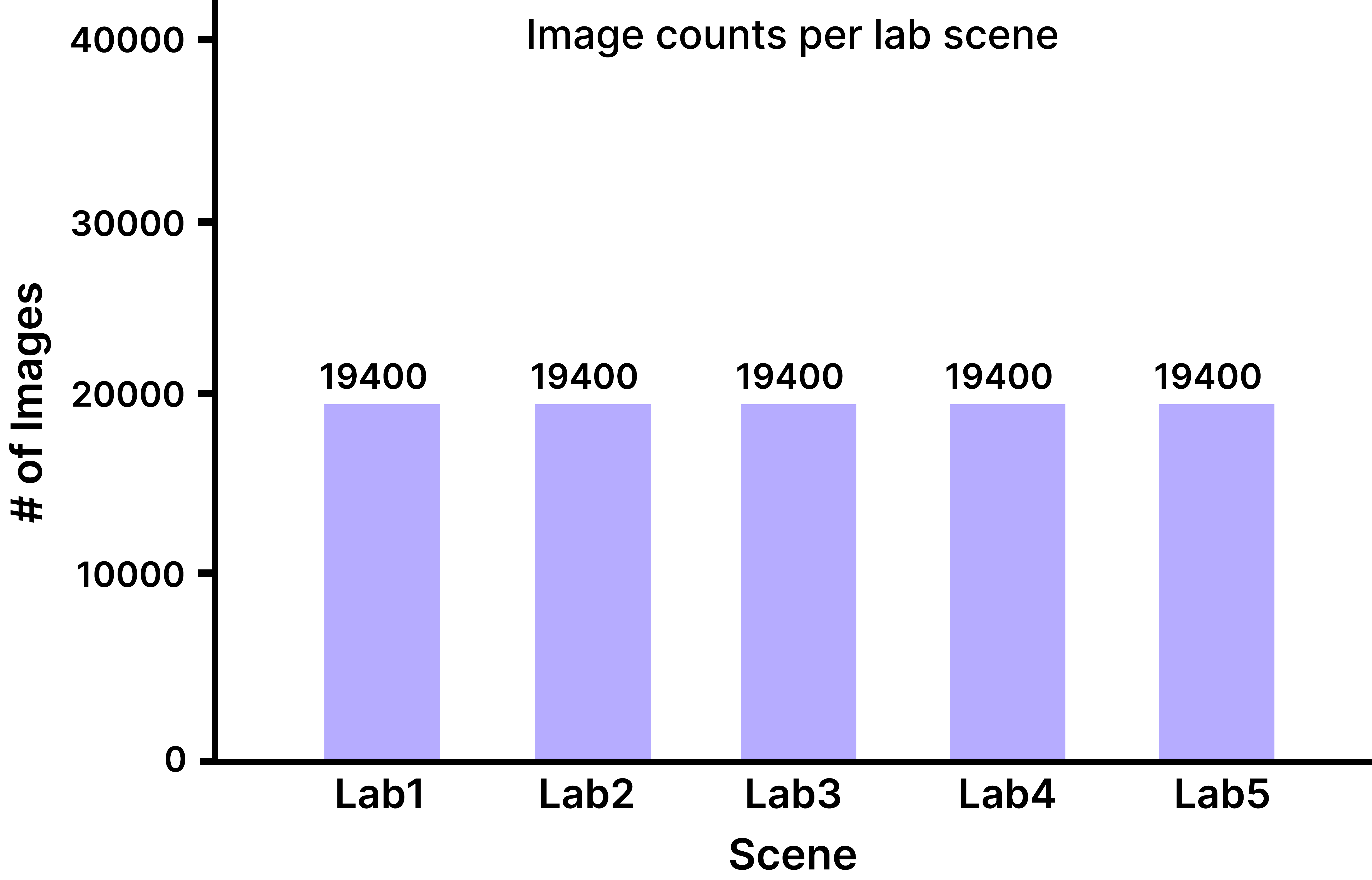}
 \caption{The number of images in each laboratory scene.}
    \label{fig:fig6c}
  \end{subfigure}
  \begin{subfigure}{0.24\linewidth}
    \includegraphics[width=1\linewidth]{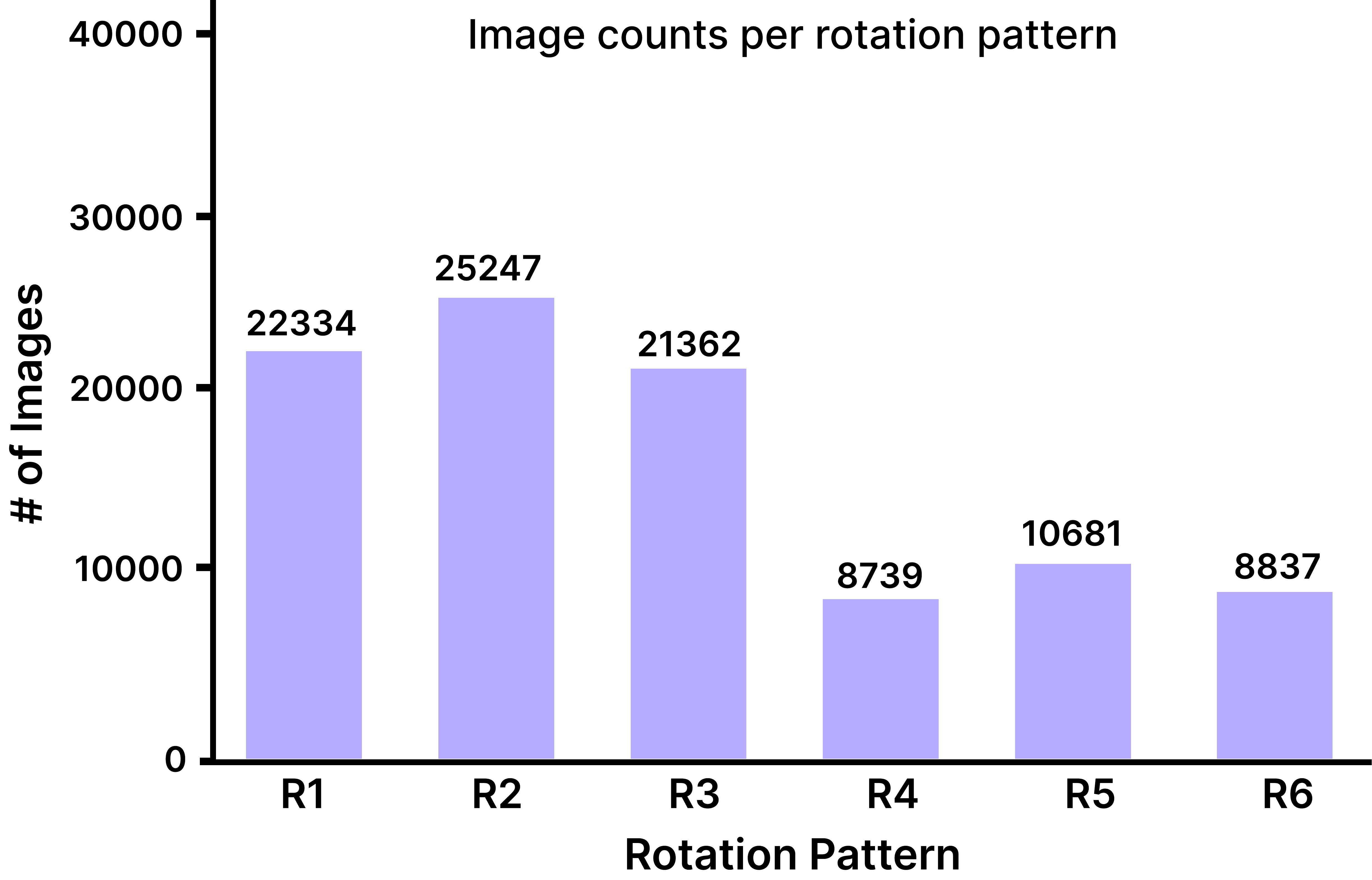}
     \caption{The number of images for each rotation mode.}
        \label{fig:fig6d}
      \end{subfigure}

\caption{Data visualization of Phys-Liquid.}
\label{fig:fig6}
\end{figure*}

A detailed analysis of the simulation factors is provided in Table~\ref{tab:dataset_comparison}, where we also compare Phys-Liquid with existing datasets collected in laboratory scenes. The distribution of image samples across conditional settings are visualized in Fig.~\ref{fig:fig6}. In detail, the five liquid colors are colorless, purple, red, orange, and yellow. The eight lighting conditions represent typical indoor illumination settings, encompassing variations in light intensity, direction, color, and mixing methods, labeled as L1 to L8. Three laboratory scenes are designated as Lab1 to Lab5 and six rotation modes are labeled from R1 to R6 (details in the appendix).

Each dataset image includes extensive metadata: container (CAD model, material, transparency), camera viewpoint (orthographic view, distance, focus), liquid properties (color, volume, mask, mesh), environmental settings (lighting, tabletop textures), physical rotation information (rotation angles, rotation mode), and image resolution.

\begin{table*}[t]
\centering
\resizebox{\textwidth}{!}{%
\begin{tabular}{l|c|c|c|c|c|c|c|c|c}
\hline
Dataset         & \# objects & \# scenes & \# images & \# liquid color & \# lighting condition & liquid deformation & multiple viewpoints &  temporal changes  & continuous volume \\ \hline
Gautham et al.\cite{narasimhan2022self}     & 2          & 1         & 4,601     & 2               & 1                     & $\times$       & $\times$    & $\times$          & $\checkmark$         \\ 
DTLD\cite{wang2024towards}      & 4          & 3         & 27,458    & 5               & 7                     & $\times$   & $\checkmark$    & $\times$      & $\checkmark$      \\ 
Phys-Liquid(Ours)    &      20      &      5     &     97,200      &  5  &    8      &    $\checkmark$                   &  $\checkmark$              &       $\checkmark$    &       $\checkmark$                           \\ \hline
\end{tabular}%
}
\caption{The analysis of Phys-Liquid dataset compared with related liquid datasets.}
\label{tab:dataset_comparison}
\end{table*}

\section{Liquid Reconstruction Pipeline}
\label{sec:method}

\begin{figure*}[h]
    \centering
    \includegraphics[width=1\textwidth]{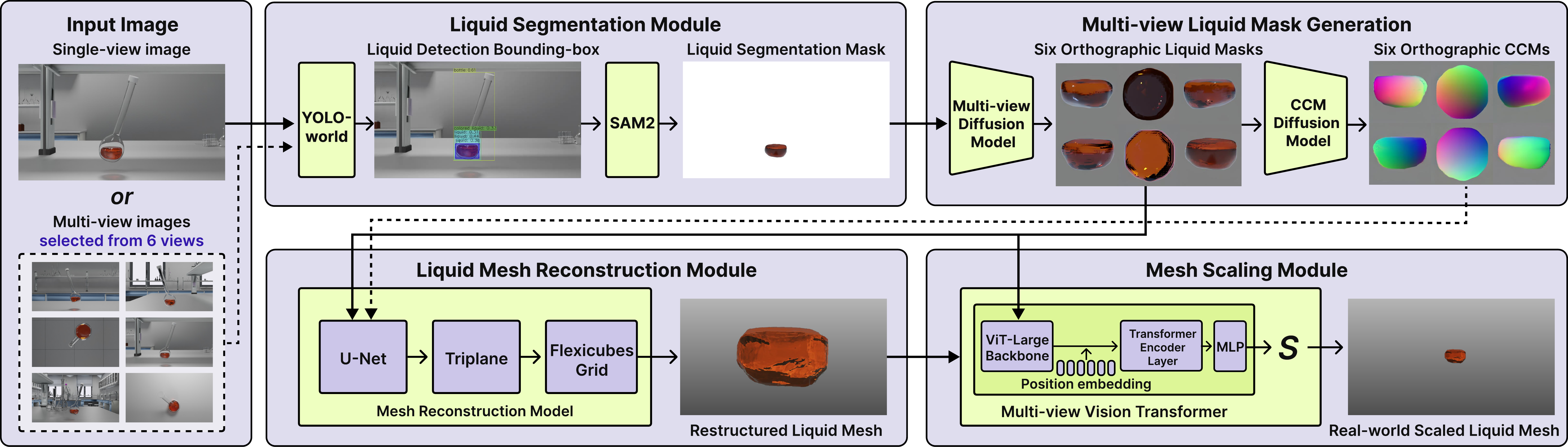}
    \caption{Overview of our four-step pipeline for reconstructing and scaling 3D meshes of deformable liquids. The input module accepts single or multi-view images, conditioned on the laboratory setup. Output examples for each step are shown.}
    \label{fig:fig7}
\end{figure*}

\subsection{Task Formulation}
Our goal is to reconstruct the 3D mesh \( S \) of deformable, transparent liquids in real-world dimensions from a single input image \( I \), formulated as:
\begin{equation}
S = F(I) = T(R(G(S(I))), s)
    \label{eq:important}
\end{equation}
We first extract a high-confidence liquid mask \( M = S(I) \) from the image \( I \) using segmentation; then, we generate six orthographic liquid masks \( \{ M_i \}_{i=1,2,\dots,6} = G(M) \) via a diffusion model to simulate multiple views. Next, we reconstruct the 3D mesh \( V = R(\{ M_i \}_{i=1,2,\dots,6}) \) by integrating the multi-view masks using a triplane method; finally, we scale the reconstructed mesh to real-world dimensions \( S = T(V, s) \), where \( s \) is a scaling factor. The pipeline framework is illustrated in Figure~\ref{fig:fig7}.

\subsection{Liquid Segmentation} 
\label{subsec:4_2}
We use SAM2~\cite{ravi2024sam} to segment liquid regions from the input RGB image. To enhance segmentation accuracy, we integrate YOLO-world~\cite{cheng2024yolo}, a real-time detector, by specifying "liquid" and "colored liquid" as positive classes and explicitly excluding the "bottle" class. A high-confidence bounding box from YOLO-world guides SAM2, resulting in precise liquid masks. As shown in the input module of Figure~\ref{fig:fig7}, the pipeline can also take multi-view images as input where segmented masks from these images are fed to the multi-view generation module.

\subsection{Multi-View Liquid Mask Generation} 
\label{subsec:4_3}
Generating accurate multi-view liquid masks is challenging for pre-trained multi-view diffusion models due to complex physics-driven deformations. We utilize the multi-view diffusion model CRM~\cite{wang2025crm}, fine-tuned on our Phys-Liquid dataset. Our approach leverages the dataset’s six orthographic views per timestep during fine-tuning. Canonical coordinate maps (CCMs) generated alongside these masks capture spatial consistency and deformation details. During generation, the fine-tuned model uses single or multi-view masks from the segmentation module to predict and complete masks from other views.  

\subsection{3D Mesh Reconstruction}
\label{subsec:4_4}
We reconstruct the 3D liquid mesh from multi-view masks and CCMs using the convolutional reconstruction model from CRM~\cite{wang2025crm}. This model employs a triplane representation, aggregating spatial features from three orthogonal planes ($xy$, $xz$, $yz$). A convolutional U-Net encodes multi-view masks into triplane features, which multi-layer perceptrons (MLPs)~\cite{rumelhart1986learning} subsequently decode into a textured 3D mesh. Due to the high fidelity of the input multi-view images, no additional fine-tuning on Phys-Liquid is necessary.

\subsection{Scaling to Real-World Dimensions} 
\label{subsec:4_5}
To align the reconstructed 3D mesh with real-world dimensions, we develop a mesh scaling model that enables the pipeline to finally estimate the geometric and volumetric properties of transparent deformable liquids in laboratory settings. The scaling model utilizes a multi-view Vision Transformer (ViT)~\cite{dosovitskiy2020image} architecture depicted in Figure~\ref{fig:fig7}. Six orthographic views are encoded separately via the ViT backbone, then integrated using positional encoding and transformer encoder layers. The combined features are processed by an MLP\cite{rumelhart1986learning} to regress a scaling factor $s$. The model is supervised using an L2 loss, with ground-truth scaling factors provided by Phys-Liquid. $s$ is calculated by comparing reconstructed dimensions with the real-world dimensions from Phys-Liquid $S_{\text{PI}}$ along the $x$, $y$, and $z$ axes:
\begin{equation}
s = \sqrt[3]{\frac{S_{\text{PI},x}}{V_x} \cdot \frac{S_{\text{PI},y}}{V_y} \cdot \frac{S_{\text{PI},z}}{V_z}}
\end{equation}

\section{Experiment}
\label{sec:experiment}

We performed two experiments
comparing our method with baselines, four experiments
evaluating the pipeline on real-world generalization, multi-view consistency, fine-tuning impact, and temporal consistency, and one ablation study on pipeline modules. The multi-view diffusion model was fine-tuned on two RTX 6000 Ada 48GB GPUs for 16 hours over 10k iterations and the scaling model trained on one RTX 6000 Ada 48GB GPU for 12 hours across 500 iterations.

\subsection{Dataset}

We used the Phys-Liquid dataset consisting of 100 separate temporal sequences, splitting it into training and testing sets (9:1) by entire sequences. Each sequence contains 81 timesteps with multi-view images. Entire sequences (all 81 timesteps within one sequence) are strictly assigned to either the training or testing set, ensuring no overlap between consecutive frames from the same sequence across sets.

\subsection{Evaluation Metrics}

For 2D mask accuracy, we report Intersection over Union (IoU). To assess 3D mesh similarity, we use Chamfer Distance (CD), Volumn IoU, and F-Score (following One-2-3-45~\cite{liu2023one}). Additionally, we compute the Root Mean Squared Error (RMSE) for the reconstructed mesh’s length, width, and height, and the Mean Absolute Percentage Error (MAPE) for the estimated scaling factor. Precise metric definitions are provided in the appendix.








\subsection{Comparison with Liquid-Specific Baseline}
\label{sec:liquidbaseline}

We compared our pipeline against a liquid-specific baseline by Eppel et al.~\cite{eppel2022predicting}, which predicts XYZ maps of transparent liquids. Their model was trained on Phys-Liquid dataset, and predictions were converted into meshes for quantitative evaluation using RMSE, Chamfer Distance, Volume IoU, and F-score on test set. Results are shown in Table~\ref{tab:liquidbaseline}. Our method outperformed the baseline, achieving lower RMSE (0.0192 vs. 0.0842) and Chamfer Distance, and higher Volume IoU and F-score, showing superior ability in capturing subtle liquid geometry and deformation.

\begin{table}[ht]
  \centering
  \scriptsize
  \resizebox{\columnwidth}{!}{
    \begin{tabular}{@{}lcccc@{}}
      \toprule
      Method & RMSE & Chamfer Distance & Volume IoU & F-Score (\%) \\
      \midrule
      Eppel et al.~\cite{eppel2022predicting} (test) & 0.0842 & 0.0412 & 0.1216 & 30.91 \\
      Our method with fine-tuning (test) & 0.0192 & 0.0079 & 0.4748 & 75.38 \\
      \bottomrule
    \end{tabular}
  }
  \caption{Reconstruction quality comparison between Eppel et al.~\cite{eppel2022predicting} and our method (threshold = 0.005 for F-Score).}
  \label{tab:liquidbaseline}
\end{table}

\subsection{Comparisons with Reconstruction Baselines}
\label{sec:reconbaseline}

\begin{figure}[h]
    \centering
    \includegraphics[width=0.48\textwidth]{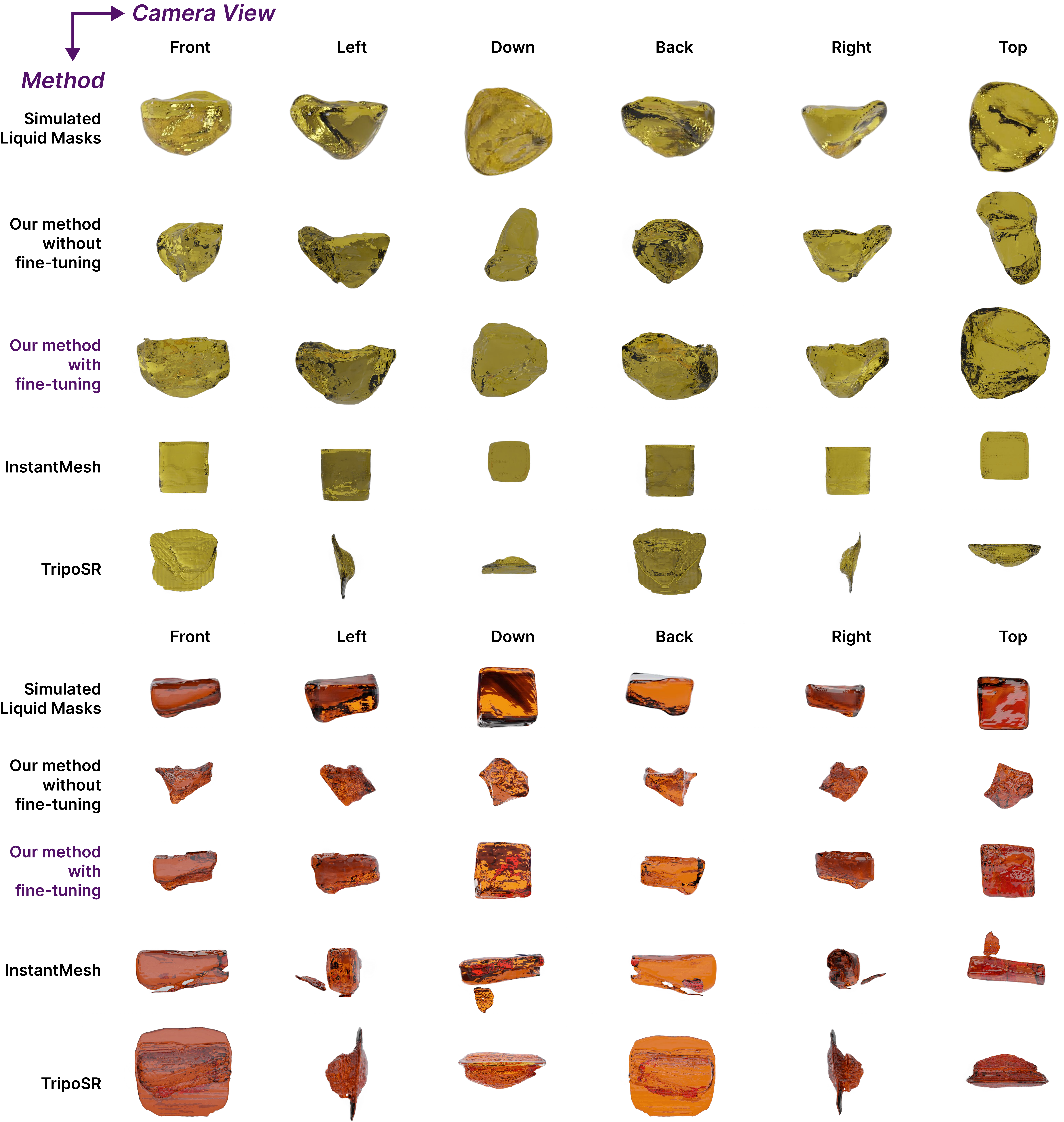}
    \caption{Qualitative comparison of reconstructed meshes by baseline methods and our pipeline with and without fine-tuning against physics-informed simulation meshes.}
    \label{fig:fig8}
\end{figure}

\paragraph{Qualitative Results} We compared the reconstruction network in our pipeline with baseline methods, including InstantMesh~\cite{xu2024instantmesh} and TripoSR~\cite{tochilkin2024triposr}. We show two example cases in the test set in Figure~\ref{fig:fig8}. We included six orthographic simulation masks from Phys-Liquid for reference, comparing these to outputs from baseline methods and our pipeline with and without fine-tuning multi-view diffusion model. The baseline methods show visual discrepancies from simulation results and fail to capture precise morphological features of liquid deformation, demonstrating the higher fidelity of our method.

\paragraph{Quantitative Results} We quantitatively compared our method with InstantMesh~\cite{xu2024instantmesh} and TripoSR~\cite{tochilkin2024triposr} using 50 randomly selected test images. Meshes were scaled to a unified bounding box for fair evaluation. Results using Chamfer Distance, Volume IoU, and F-Score are summarized in Table~\ref{tab:baseline_comparison}. Our method outperformed two baseline methods, indicating higher reconstruction accuracy.

\begin{table}[ht]
  \centering
  \scriptsize
  \resizebox{\columnwidth}{!}{
    \begin{tabular}{@{}lccc@{}}
      \toprule
      Method & Chamfer Distance & Volume IoU & F-Score (\%) \\
      \midrule
      InstantMesh~\cite{xu2024instantmesh} & 0.0189 & 0.2794 & 46.18 \\
      TripoSR~\cite{tochilkin2024triposr} & 0.0275 & 0.2275 & 38.06 \\
      Our method without fine-tuning & 0.0128 & 0.3246 & 58.19 \\
      Our method with fine-tuning & \textbf{0.0085} & \textbf{0.6236} & \textbf{78.57} \\
      \bottomrule
    \end{tabular}
  }
  \caption{Quantitative reconstruction comparison with baseline methods (threshold = 0.005) using \textbf{50} test images.}
  \label{tab:baseline_comparison}
\end{table}

\subsection{Generalization to Real-World Data}
\label{sec:realworld_comparison}

To evaluate the applicability and scalability of our method beyond simulated laboratory environments, we tested our model trained exclusively on the Phys-Liquid dataset directly on the DTLD dataset~\cite{wang2024towards}, a real-world dataset featuring static liquid states without deformation. Performance was measured and presented in Table~\ref{tab:realworld_comparison}. Although DTLD~\cite{wang2024towards} lacks liquid deformation, the results indicate that our method maintains reasonable accuracy in real-world scenarios, indicating that the physical priors and visual characteristics in Phys-Liquid are closely aligned with real-world conditions to support transferability.

\begin{table}[ht]
  \centering
  \scriptsize
  \resizebox{\columnwidth}{!}{
    \begin{tabular}{@{}lcccc@{}}
      \toprule
      Dataset & RMSE & Chamfer Distance & Volume IoU & F-Score (\%) \\
      \midrule
      DTLD dataset~\cite{wang2024towards} & 0.0266 & 0.0172 & 0.3861 & 62.43 \\
      Phys-Liquid dataset (test) & 0.0192 & 0.0079 & 0.4748 & 75.38 \\
      \bottomrule
    \end{tabular}
  }
  \caption{Quantitative comparison of reconstruction performance on DTLD~\cite{wang2024towards} and Phys-Liquid datasets (threshold = 0.005) using \textbf{entire} test set.}
  \label{tab:realworld_comparison}
\end{table}

\subsection{Multi-View Consistency of Mask Generation}
\label{sec:5_5}

We evaluated fine-tuning the diffusion model on the Phys-Liquid dataset for multi-view mask generation. Performance was assessed on the test set by comparing generated masks with simulated masks using IoU and visual inspection. Fine-tuning improved average IoU from 74.38\% to 90.05\%, yielding masks with more realistic liquid boundaries. Figure \ref{fig:fig8} presents a visual comparison, validating the dataset's effectiveness in improving generation precision.

\begin{figure}[h]
    \centering
    \includegraphics[width=0.3\textwidth]{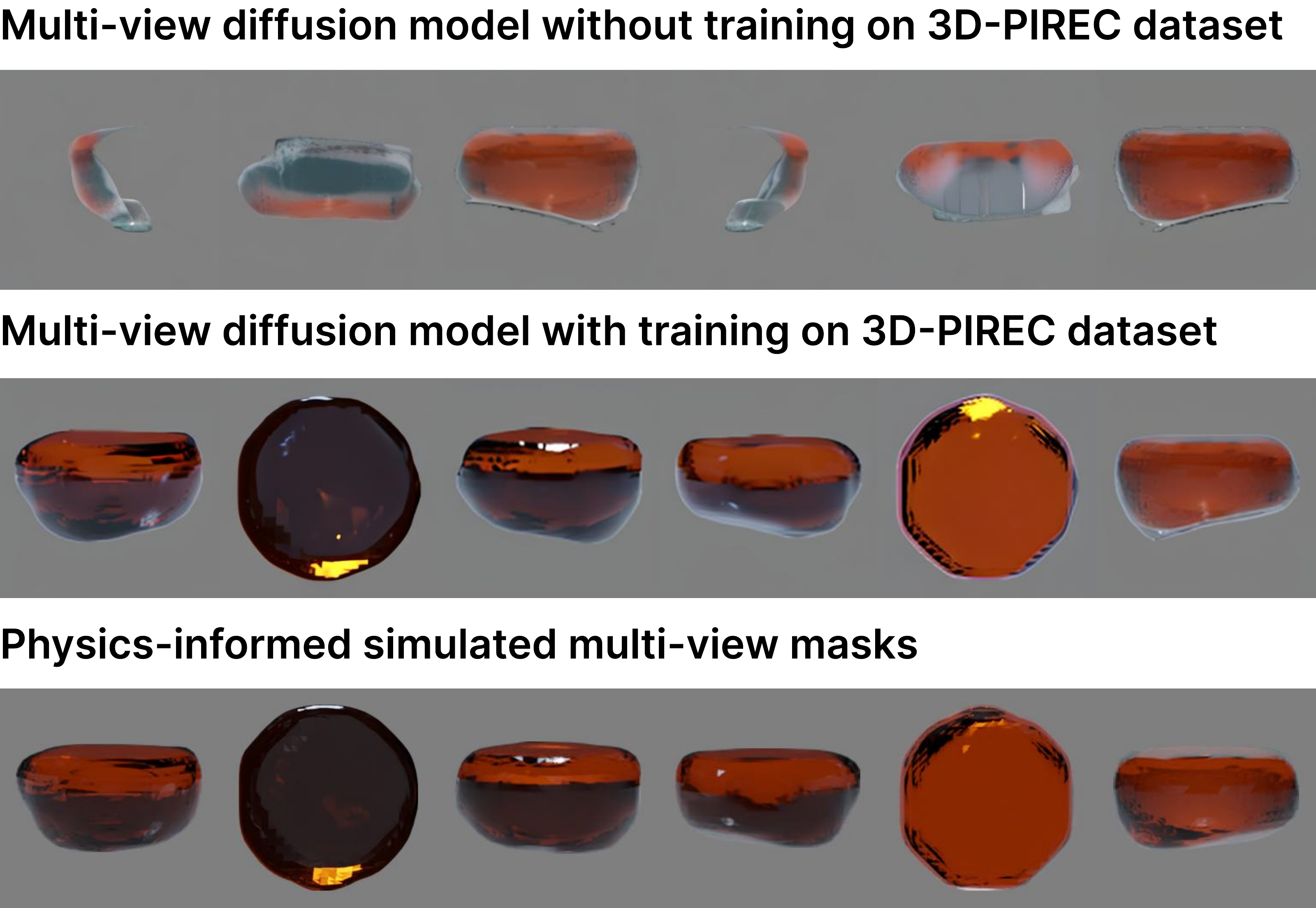}
    \caption{Multi-view liquid masks before and after diffusion model fine-tuning, compared with physics-informed masks.}
    \label{fig:fig9}
\end{figure}

We further evaluated multi-view consistency using different single-view inputs. We computed the average IoU between predicted multi-view masks and simulated masks across the test set for each camera view (Table~\ref{tab:multi_view_consistency}). Results confirm high consistency across multi-view predictions.

\begin{table}[ht]
  \centering
  \scriptsize
  \resizebox{\columnwidth}{!}{
    \begin{tabular}{@{}lcccccc@{}}
      \toprule
      Camera View & Front & Left & Down & Back & Right & Top \\
      \midrule
      Average IoU (\%) & 91.76 & 90.16 & 89.32 & 91.03 & 90.82 & 89.21 \\
      \bottomrule
  \end{tabular}}
  \caption{Average IoU between generated and physics-informed simulation masks across camera views.}
  \label{tab:multi_view_consistency}
\end{table}

\subsection{Diffusion Model Fine-Tuning Impact}
\label{sec:5_7}

We evaluated the impact of diffusion model fine-tuning on the overall 3D reconstruction quality. We generated multi-view masks before and after fine-tuning and reconstructed meshes using our pipeline. Reconstructions were quantitatively assessed. As shown in Table~\ref{tab:tab3}, the fine-tuned model achieved improvement across all metrics, reducing the RMSE from 2.54\% to 1.92\% on the test set. The results confirm that fine-tuning enhances reconstruction accuracy.

\begin{table}[ht]
  \centering
  \scriptsize 
  \resizebox{\columnwidth}{!}{ 
    \begin{tabular}{@{}lcccc@{}}
      \toprule
      Method     & RMSE   & Chamfer Distance & Volume IoU & F-Score (\%) \\
      \midrule
      without fine-tuning (test) & 0.0254 & 0.0139        & 0.2850   & 46.19 \\
      fine-tuning (training) & 0.0170 & 0.0083 & 0.4514 & 72.33 \\
      fine-tuning (test) & 0.0192 & 0.0079 & 0.4748 & 75.38 \\
      \bottomrule
    \end{tabular}
  }
  \caption{Reconstruction quality comparison before and after fine-tuning the diffusion model (threshold = 0.005).}
  \label{tab:tab3}
\end{table}

\subsection{Temporal Consistency of Reconstruction}
\label{sec:5_8}

We evaluated the temporal consistency of our pipeline in reconstructing liquid meshes across sequential frames from the test set of the Phys-Liquid dataset. We computed the RMSE of reconstructed meshes at each timestep within individual temporal sequences, and then calculated the variance and standard deviation of RMSE values for each sequence to quantify reconstruction stability. Over 100 temporal sequences, the average variance and standard deviation of RMSE were \(0.00038038\) and \(0.00643858\). These low values indicate that our pipeline maintains high temporal consistency, capturing stable and accurate liquid deformation states across consecutive frames.

\subsection{Ablation Study on Pipeline Modules}
\label{sec:ablation1}
We evaluated the contribution of each module in our pipeline by sequentially replacing their outputs with simulation results on the Phys-Liquid test set. Results of reconstruction quality are presented in Table~\ref{tab:ablation_pipeline}, highlighting that mesh reconstruction and mesh scaling modules influence more on reconstruction performance.

\begin{table}[ht]
  \centering
  \scriptsize
  \resizebox{\columnwidth}{!}{
    \begin{tabular}{@{}lcccc@{}}
      \toprule
      Module Replaced & RMSE & Chamfer Distance & Volume IoU & F-Score (\%) \\
      \midrule
      Segmentation & 0.0130 & 0.0075 & 0.5504 & 78.42 \\
      Multi-view Mask Generation & 0.0105 & 0.0067 & 0.6532 & 81.36 \\
      Mesh Reconstruction & 0.0085 & 0.0058 & 0.7687 & 85.64 \\
      Mesh Scaling & 0.0071 & 0.0042 & 0.7511 & 88.47 \\
      \midrule
      Entire pipeline (test) & 0.0192 & 0.0079 & 0.4748 & 75.38 \\
      \bottomrule
  \end{tabular}}
  \caption{Ablation study evaluating contributions of pipeline modules by substituting their outputs with simulation data.}
  \label{tab:ablation_pipeline}
\end{table}

\section{Conclusions and Limitations}
\label{sec:conclusions}

We introduced Phys-Liquid, a physics-informed dataset designed specifically to support research on transparent deformable liquid perception. By accurately modeling liquid dynamics under realistic laboratory conditions, our dataset addresses the critical gap left by existing static or simplified datasets. In particular, it captures 3D deformations of liquids under temporally evolving container rotations, expanding the problem space from 3D to spatiotemporal (4D) domain and enabling the study of dynamic fluid perception.

While modest in size, the dataset emphasizes physically grounded diversity, with variations in lighting, liquid types, container shapes, and motion profiles. A reconstruction pipeline demonstrates the dataset’s validity, supported by evaluations on both synthetic and real-world benchmarks.

Built within Blender, Phys-Liquid also enables the rendering of additional optical modalities—such as surface normals and refractive flow—that are difficult to obtain in real settings. This flexibility makes Phys-Liquid not only a benchmark, but also a versatile tool for training and evaluating physically grounded perception systems. We believe it provides a foundation for further research into multi-modal fluid representation and physics-aware visual reasoning.


\section*{Acknowledgments}
\label{sec:Acknowledgments}
This work was supported by the National Natural Science Foundation of China under Grant 62525604, ``Cell Therapy Designed by Artificial Intelligence''. This work was also supported by the National Natural Science Foundation of China under Grant 62271221.

\bibliography{aaai2026}


\section*{Appendix}

The appendix is organized as follows:

\begin{itemize}
    \item providing data visualization of the Phys-Liquid dataset, including image distributions across various settings and additional high-resolution image samples.

    \item presenting the results of five supplementary experiments and one ablation study on loss function.
\end{itemize}

\section{Data Visualization of the Phys-Liquid Dataset}
\label{sec:datasetSup}

We present a supplementary specification of the Phys-Liquid dataset to provide a more comprehensive understanding of the image distributions across scene settings through separated visualization charts included in the main paper. Four bar charts display the distribution of images under different scene settings in the Phys-Liquid dataset:

\begin{itemize}
    \item Figure~\ref{fig:fig9} shows the image counts for each liquid color.
    \item Figure~\ref{fig:fig10} shows the image counts under each lighting set.
    \item Figure~\ref{fig:fig11} displays the image counts in each scene.
    \item Figure~\ref{fig:fig12} illustrates the image counts corresponding to each rotation mode.
\end{itemize}

\begin{figure}[h]
    \centering
    \includegraphics[width=0.43\textwidth]{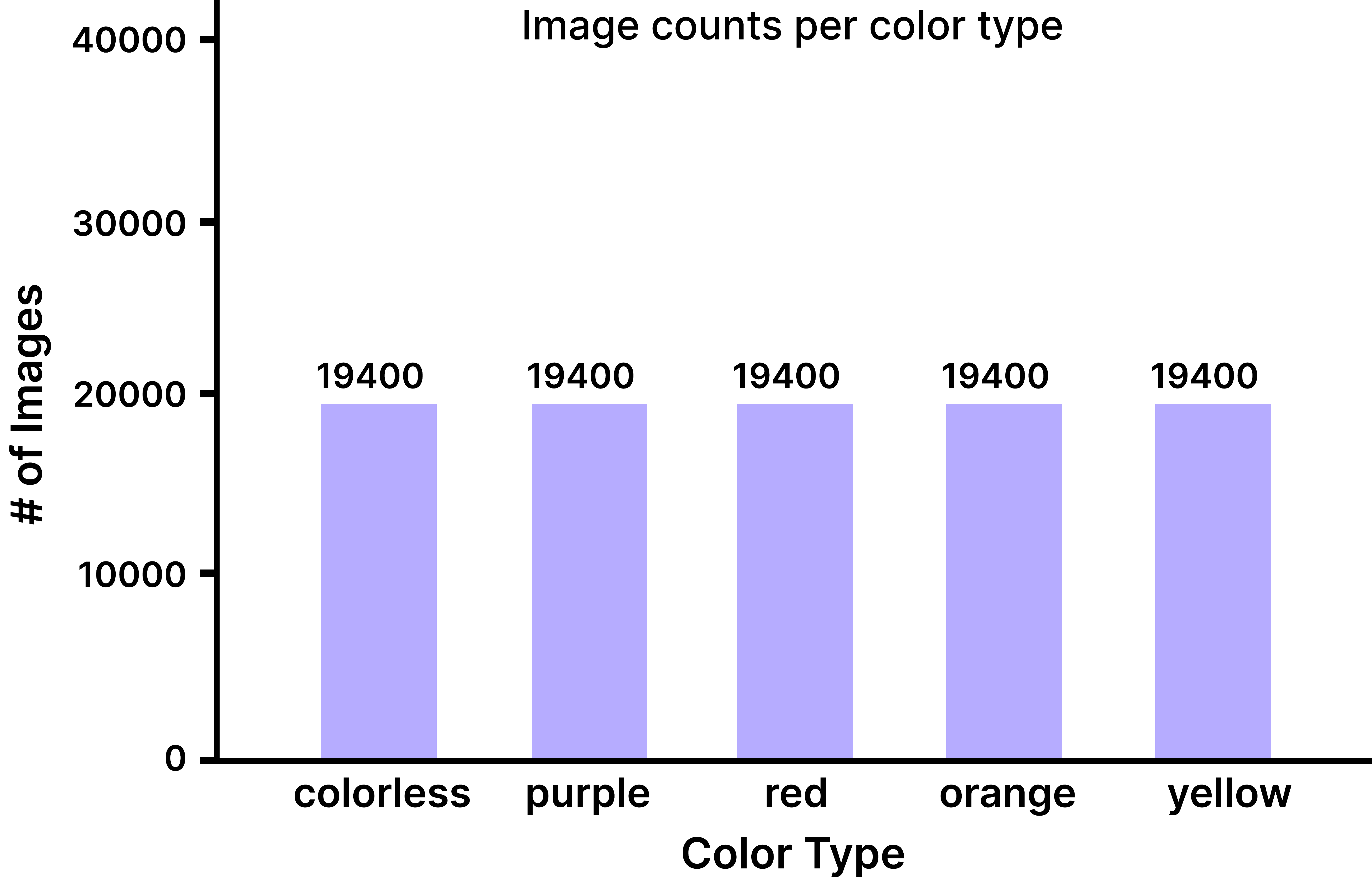}
    \caption{The number of images for each liquid color.}
    \label{fig:fig9}
\end{figure}

\begin{figure}[h]
    \centering
    \includegraphics[width=0.43\textwidth]{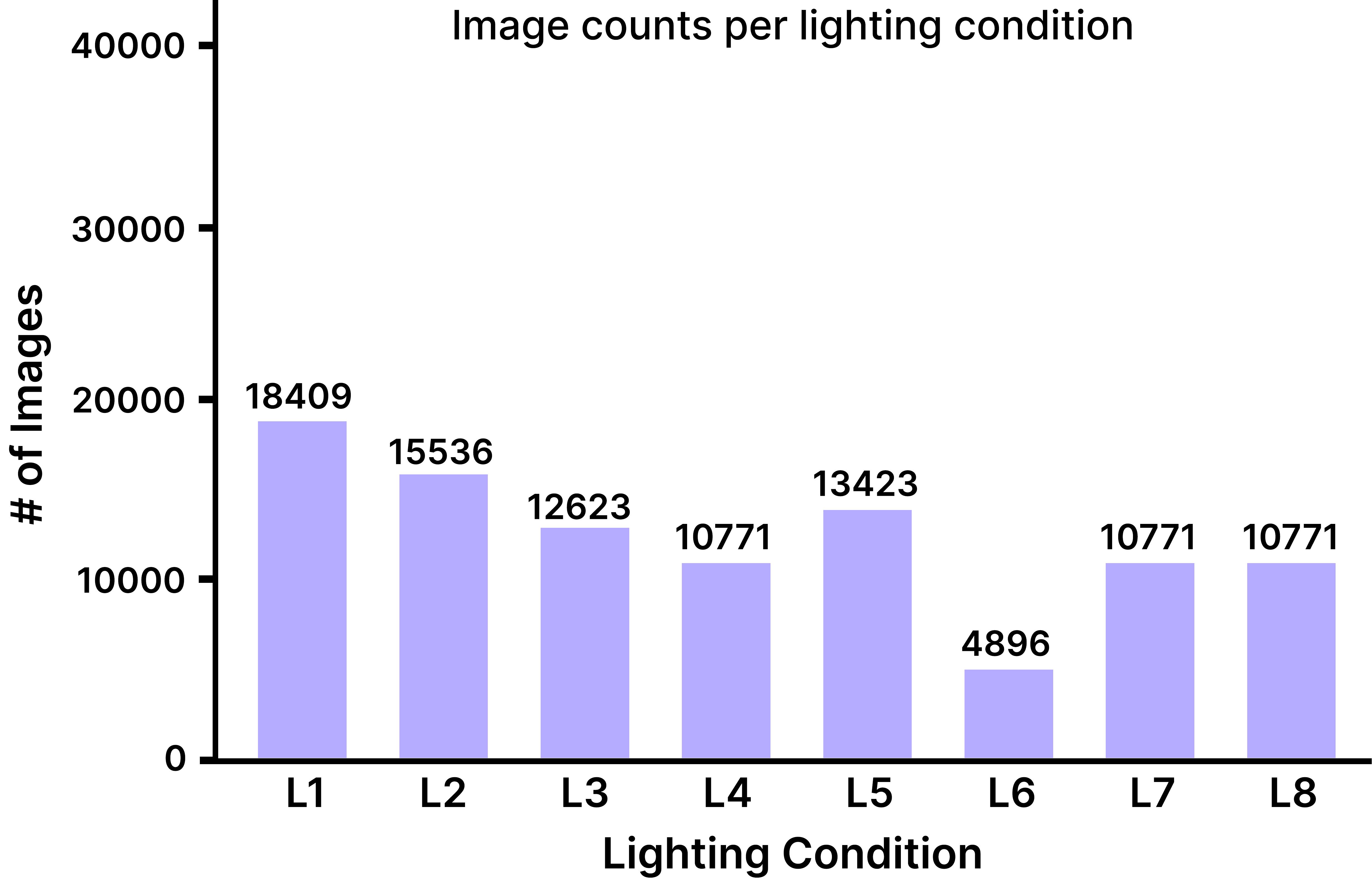}
    \caption{The number of images under each lighting condition.}
    \label{fig:fig10}
\end{figure}

\begin{figure}[h]
    \centering
    \includegraphics[width=0.43\textwidth]{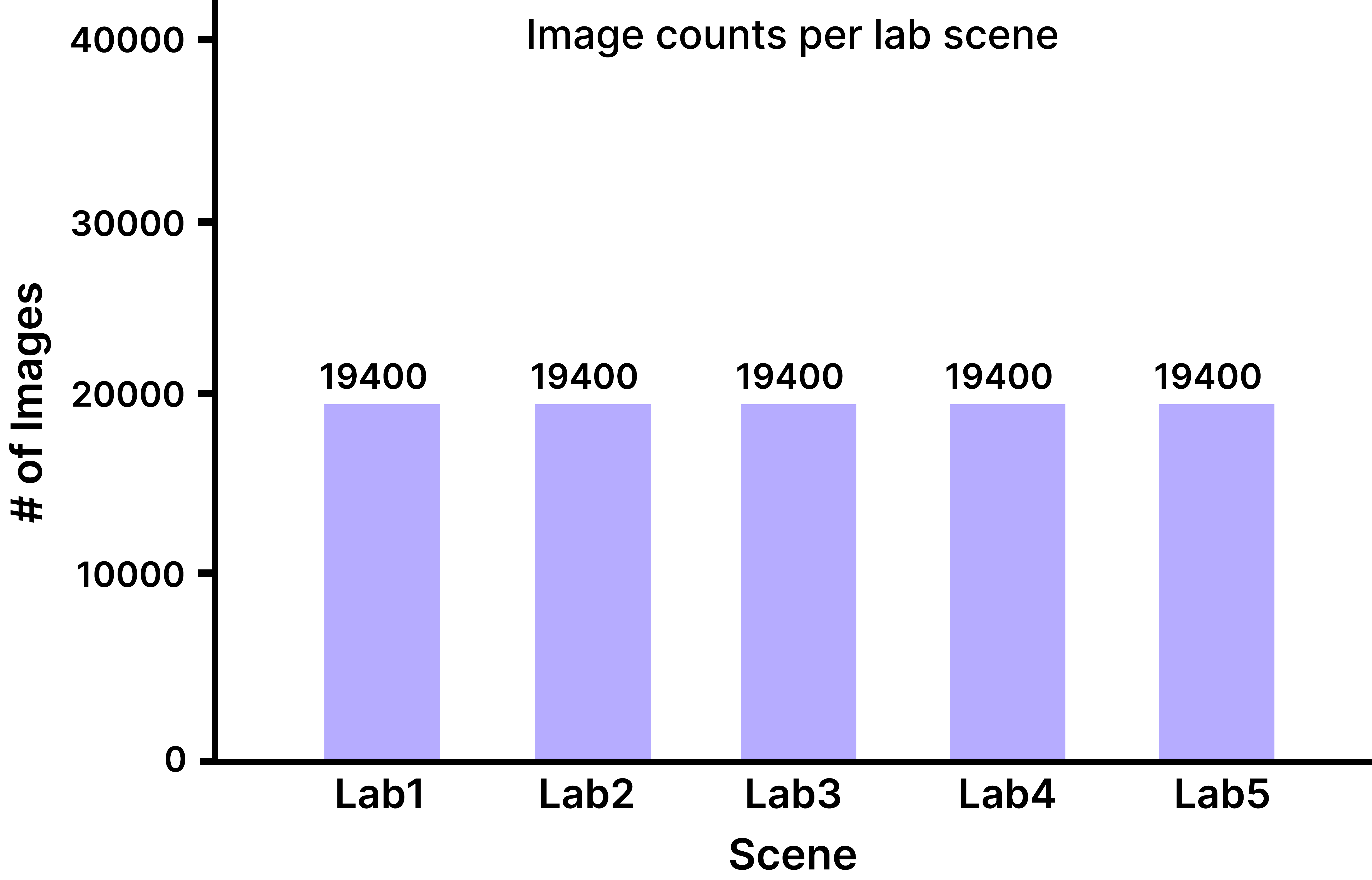}
    \caption{The number of images in each laboratory scene.}
    \label{fig:fig11}
\end{figure}

\begin{figure}[h]
    \centering
    \includegraphics[width=0.43\textwidth]{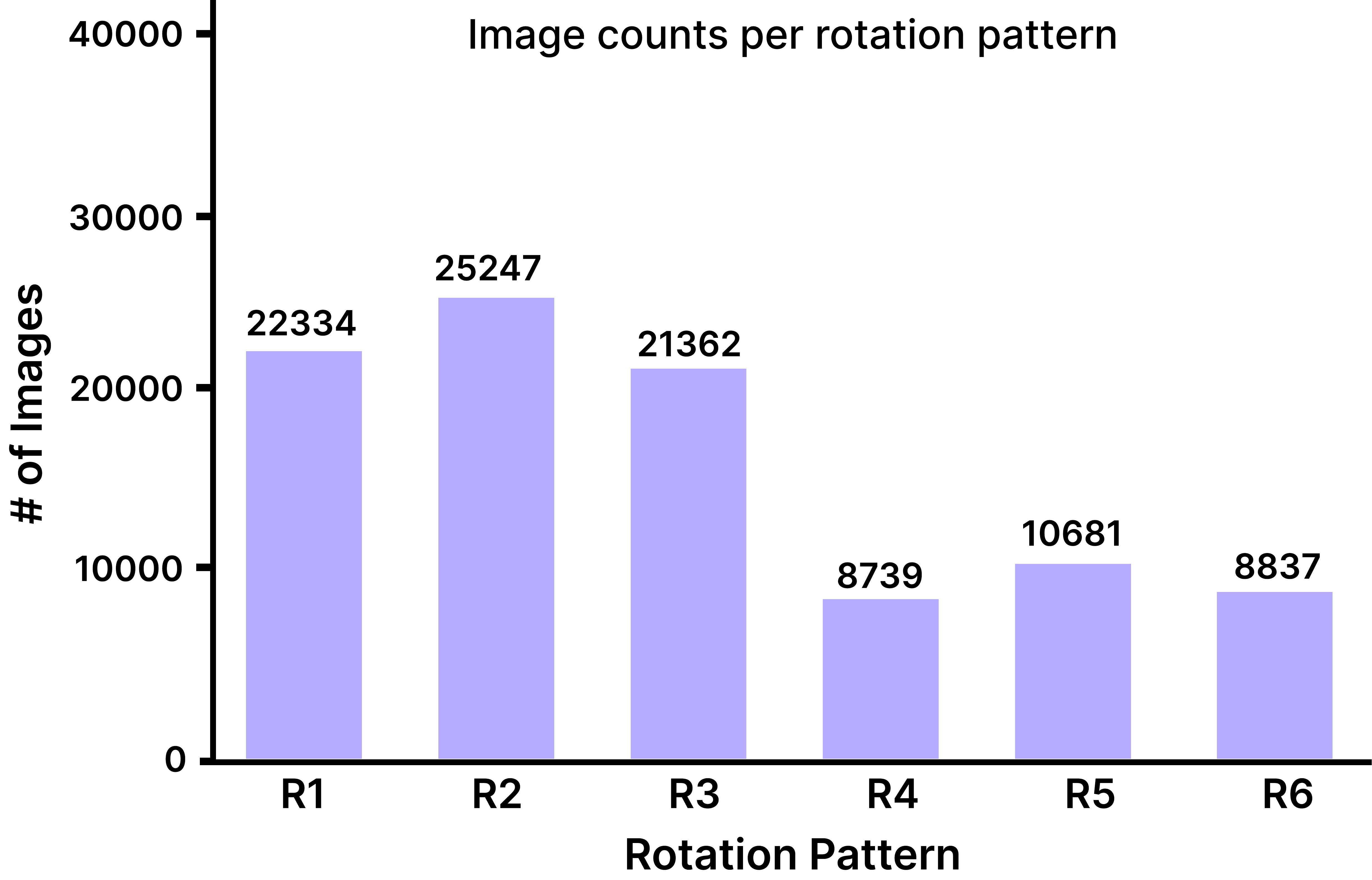}
    \caption{The number of images for each rotation mode.}
    \label{fig:fig12}
\end{figure}

\begin{figure*}[h]
    \centering
    \includegraphics[width=0.9\textwidth]{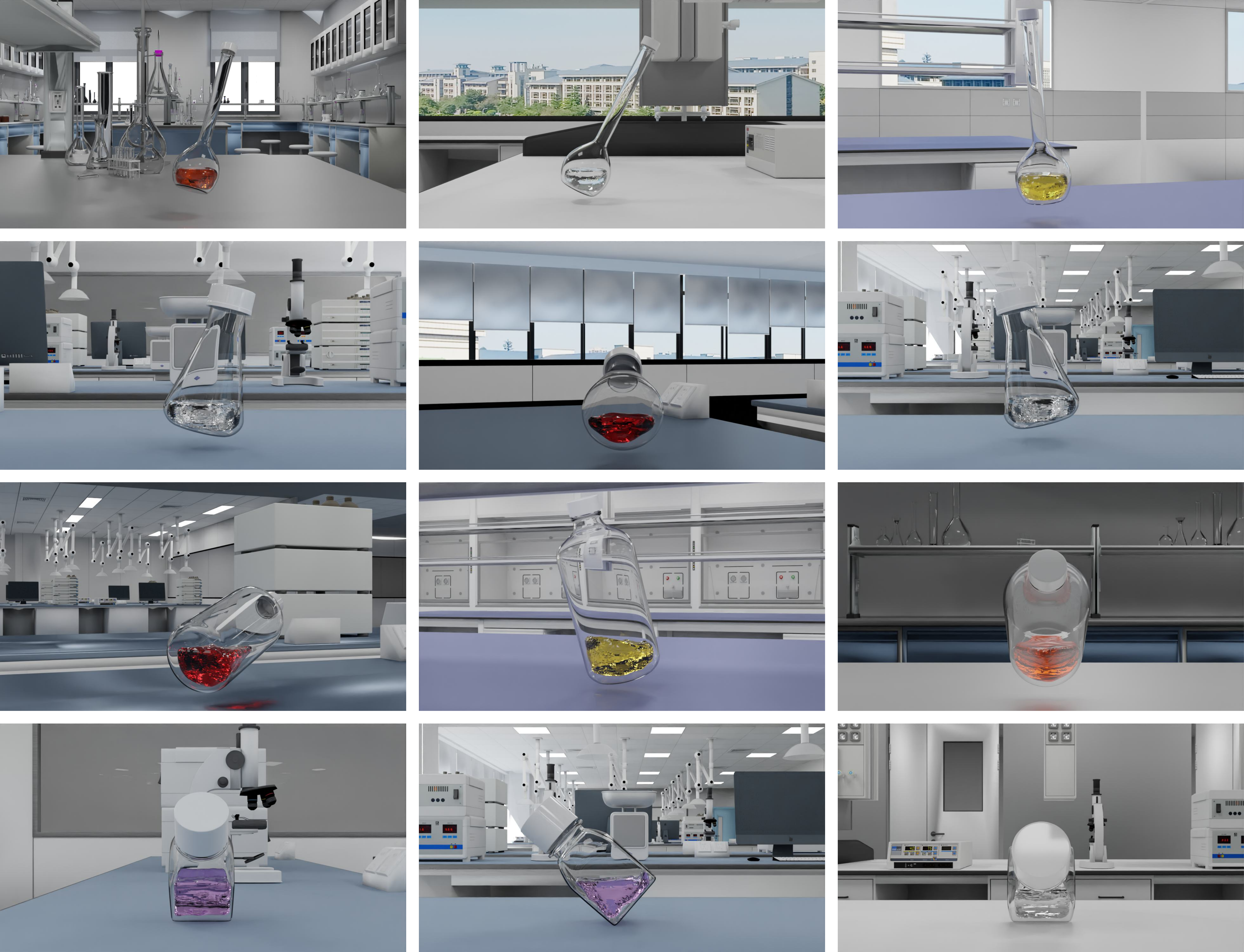}
    \caption{Additional high-resolution image samples across different scene settings.}
    \label{fig:figcases}
\end{figure*}

In detail, the five liquid colors are colorless, purple, red, orange, and yellow, which are common reagent colors used in laboratories. The eight lighting conditions represent typical indoor illumination settings, encompassing variations in light intensity, direction, color, and mixing methods, labeled as L1 to L8. The three laboratory scenes are designated as Lab1 to Lab5. The six rotation modes, labeled R1 to R6, are defined as follows:

\begin{itemize}
    \item R1: Rotation only around the X-axis from $0^\circ$ to $80^\circ$.
    \item R2: Simultaneous rotation around the X and Y axes from $0^\circ$ to $80^\circ$.
    \item R3: Rotation around the X-axis ($0^\circ$ to $40^\circ$), Y-axis ($0^\circ$ to $50^\circ$), and Z-axis ($0^\circ$ to $80^\circ$).
    \item R4: Rotation around the X-axis ($0^\circ$ to $30^\circ$), Y-axis ($0^\circ$ to $40^\circ$), and Z-axis ($0^\circ$ to $60^\circ$).
    \item R5: Rotation around the X-axis ($0^\circ$ to $80^\circ$), Y-axis ($0^\circ$ to $40^\circ$), and Z-axis ($0^\circ$ to $20^\circ$).
    \item R6: Rotation around the Y-axis ($0^\circ$ to $80^\circ$) and Z-axis ($0^\circ$ to $60^\circ$).
\end{itemize}

Each image in the dataset is accompanied by comprehensive metadata that provides detailed descriptions of various aspects of the simulation:

- \textbf{Container Object}: Includes the container name, its CAD model, material, thickness, and transparency level.
  
- \textbf{Camera Viewpoint}: Specifies one of the six orthographic views, camera focus settings, the distance from the camera to the object's center, and the camera size.
  
- \textbf{Liquid Object}: Contains information on the liquid color, initial volume, liquid mask, actual 3D dimensions of the liquid, and the liquid's OBJ 3D mesh file.
  
- \textbf{Environment Settings}: Describes the lighting condition, laboratory scene, and tabletop material and texture.
  
- \textbf{Physical Rotation Information}: Includes the rotation mode (one of R1 to R6) and the specific rotation angles around the X, Y, and Z axes.
  
- \textbf{Image Information}: Provides the image resolution.

Additionally, in Figure~\ref{fig:figcases} we present twelve high-resolution images samples of four containers with distinct background scenes from the Phys-Liquid dataset.

\begin{figure*}[h]
    \centering
    \includegraphics[width=0.78\textwidth]{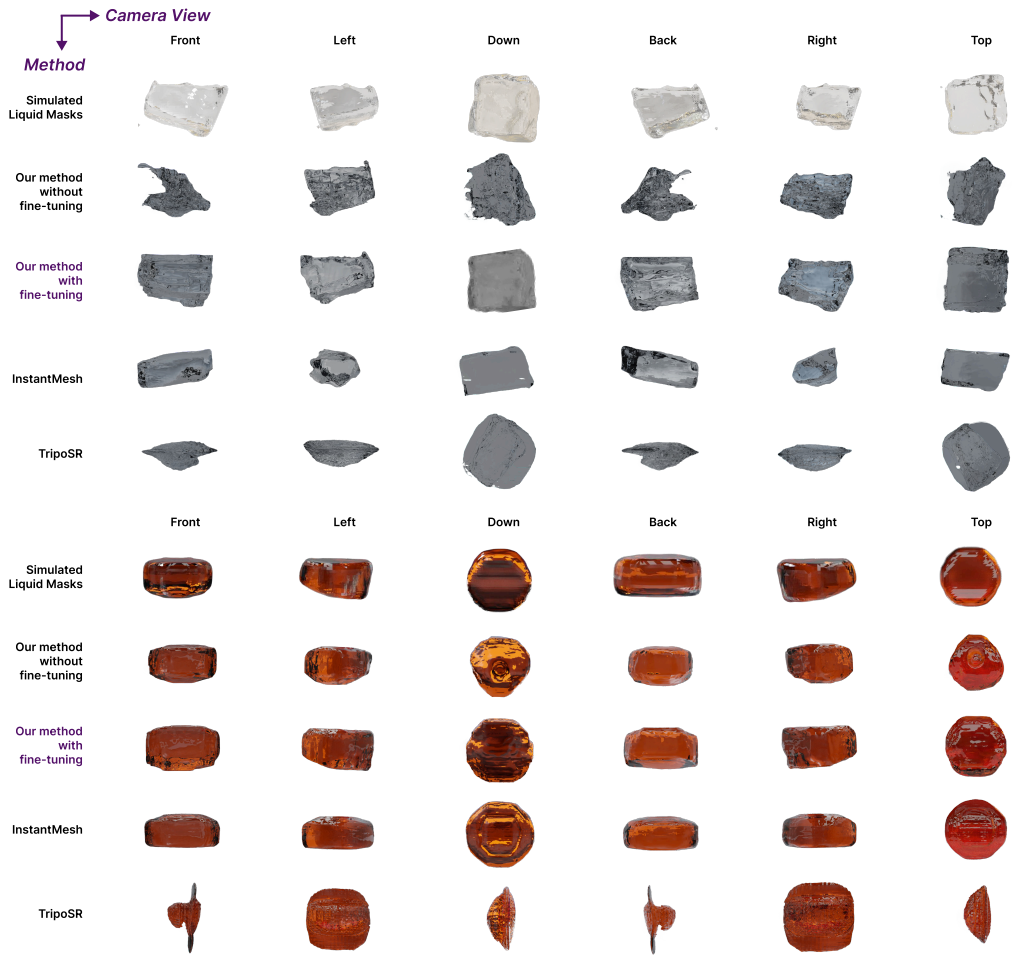}
    \caption{Comparison of 3D reconstruction results for two cases from the test set. Each case includes the simulation results from the Phys-Liquid dataset and the reconstructed outputs from InstantMesh~\cite{xu2024instantmesh}, TripoSR\cite{tochilkin2024triposr}, and our Phys-Liquid pipeline with and without fine-tuning the diffusion model.}
    \label{fig:fig13}
\end{figure*}

\begin{figure}[h]
    \centering
    \includegraphics[width=0.5\textwidth]{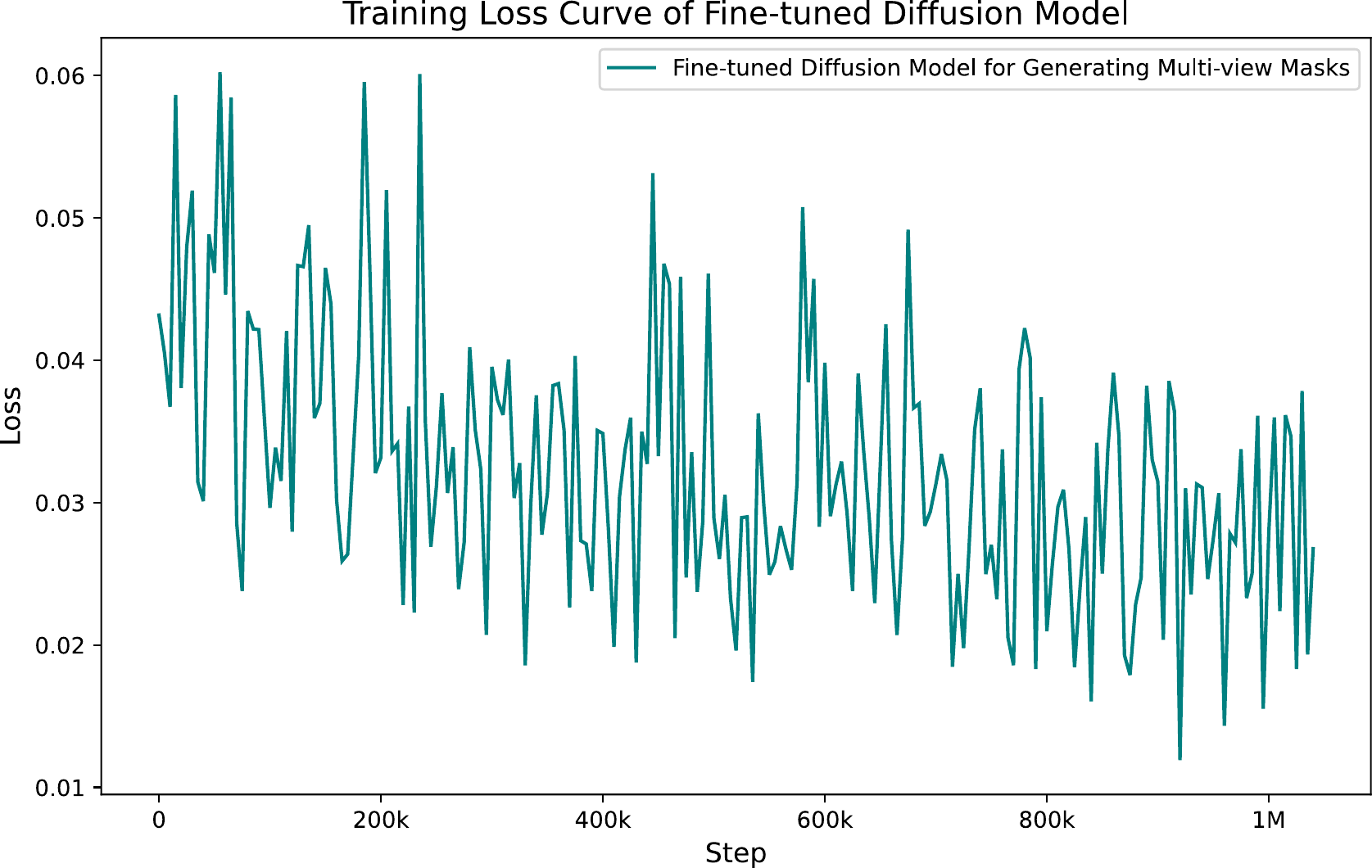}
    \caption{Loss curve during the fine-tuning of the diffusion model for generating multi-view liquid masks.}
    \label{fig:fig14}
\end{figure}

\begin{figure}[h]
    \centering
    \includegraphics[width=0.5\textwidth]{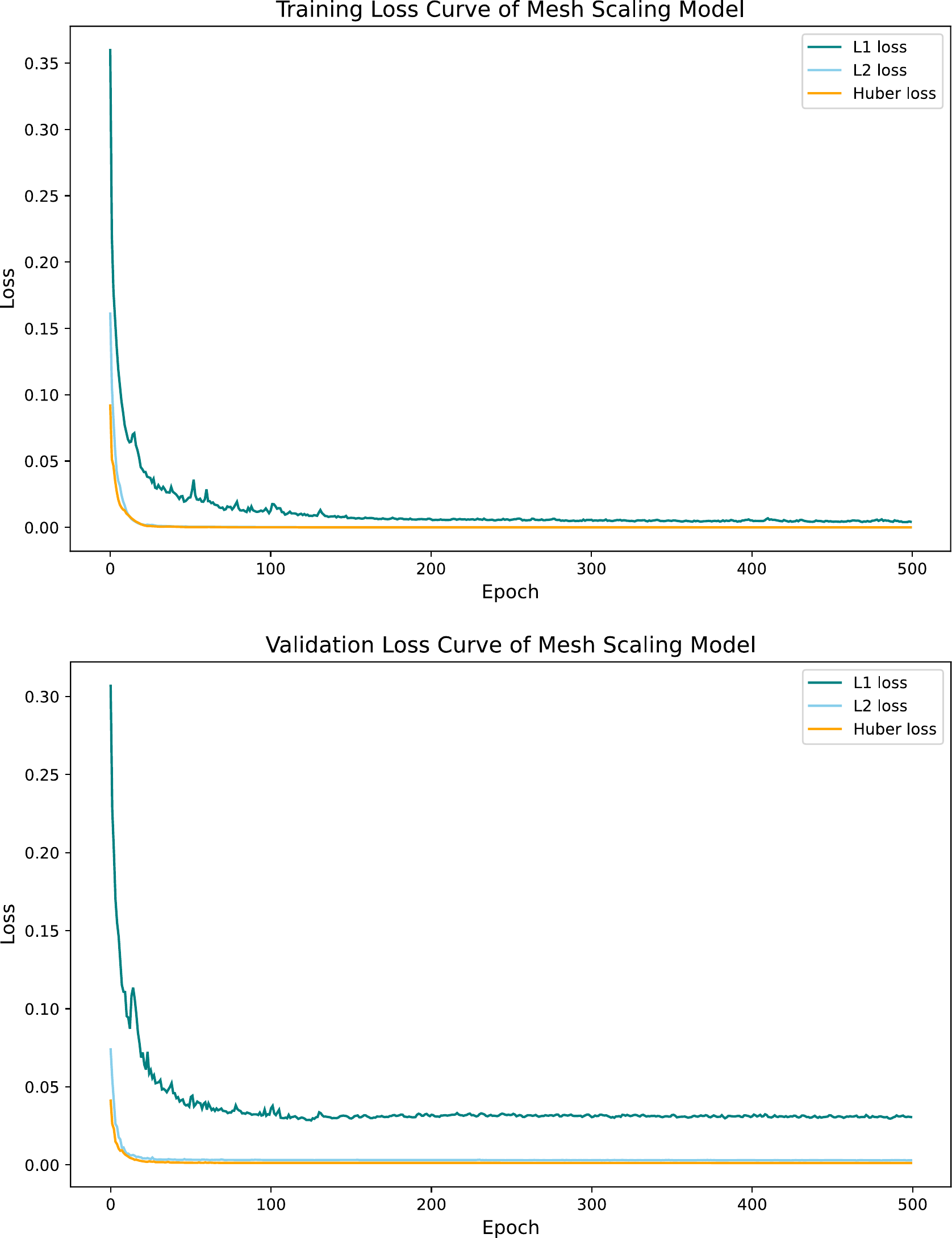}
    \caption{Training and validation loss curves of the mesh scaling model using L1 loss, L2 loss, and Huber loss.}
    \label{fig:fig15}
\end{figure}

\begin{figure}[h]
    \centering
    \includegraphics[width=0.5\textwidth]{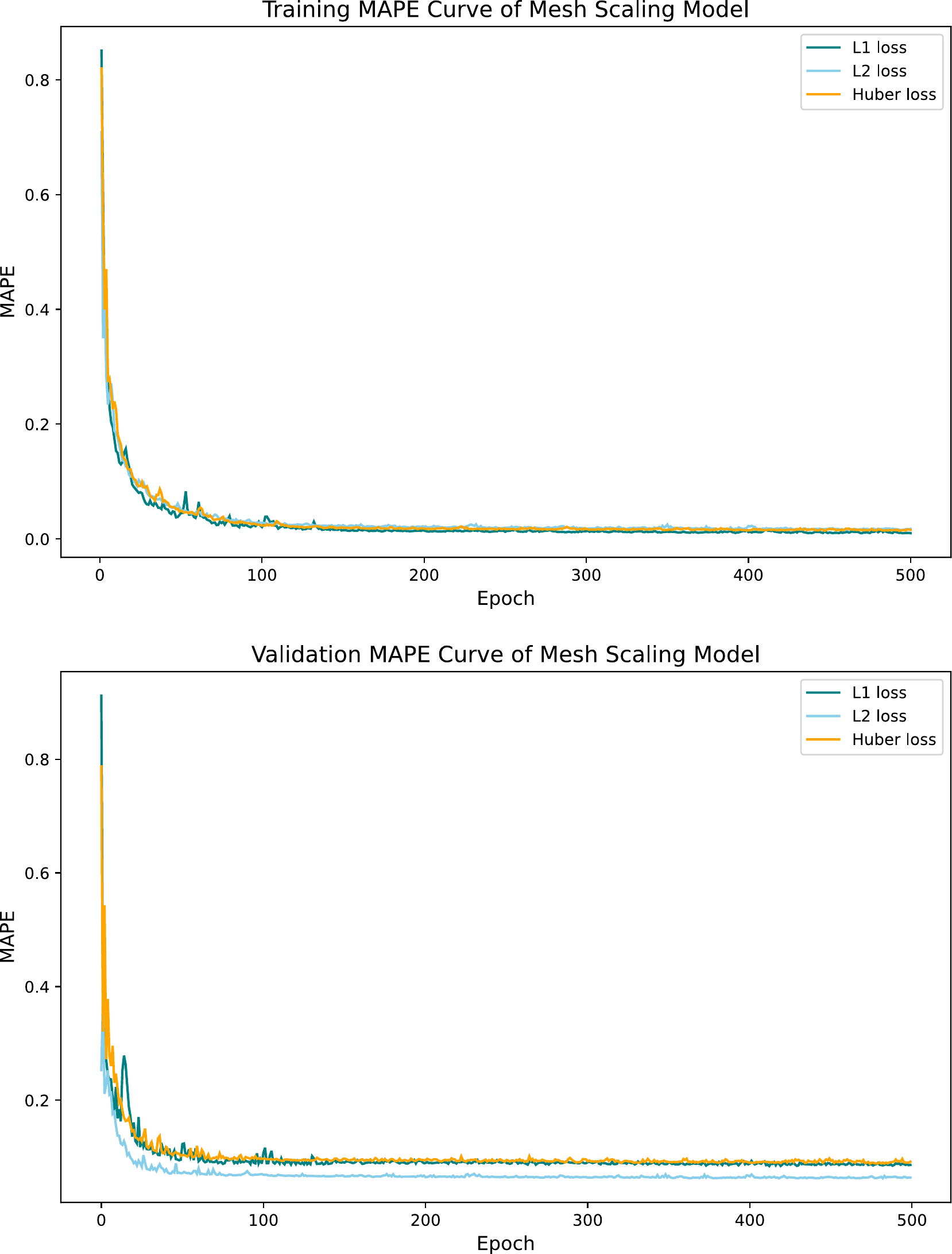}
    \caption{MAPE of the scaling factor \( s \) over 500 epochs during the training and validation of the mesh scaling model.}
    \label{fig:fig16}
\end{figure}

\begin{figure}[h]
  \centering
  \includegraphics[width=1\linewidth]{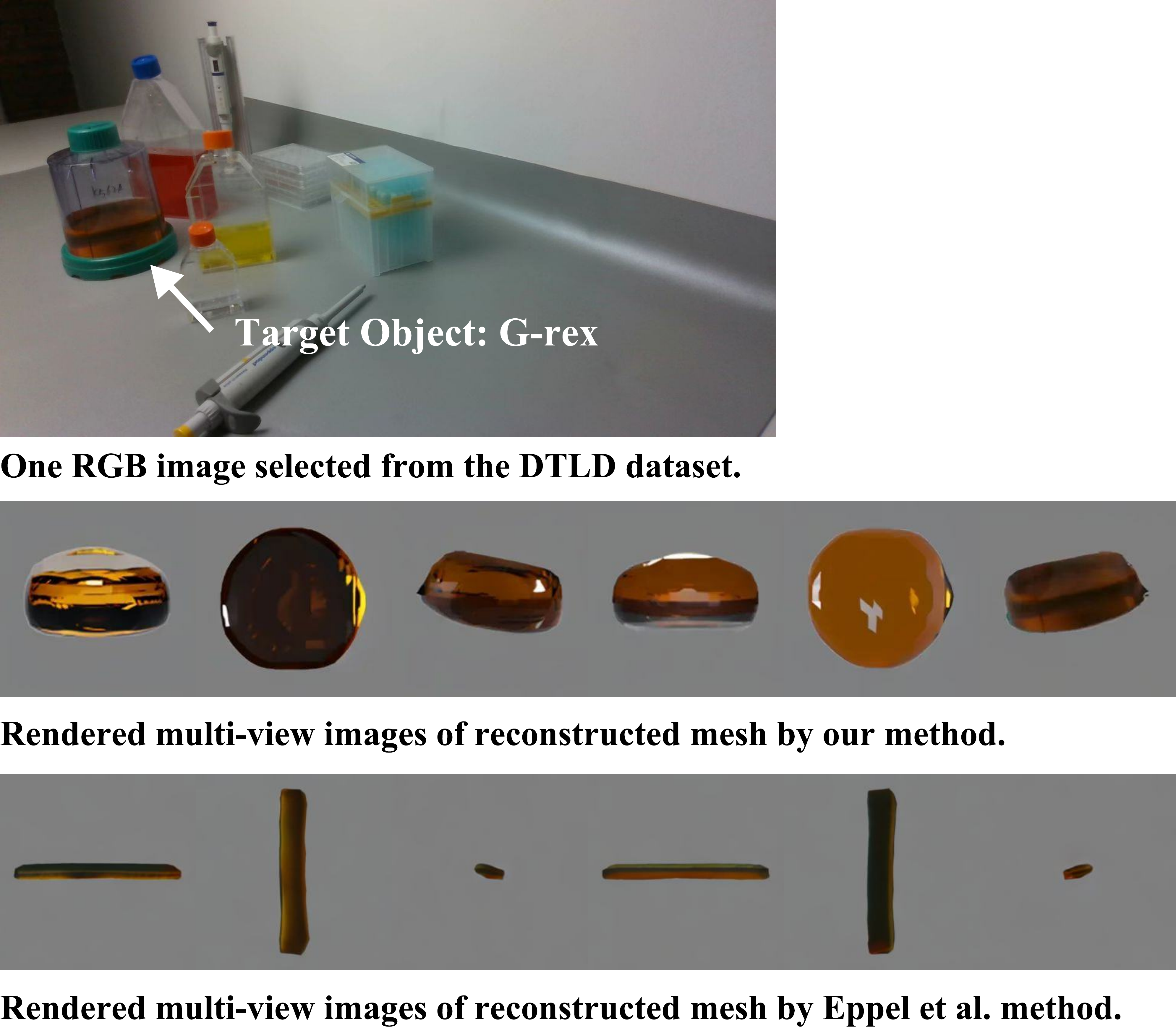}
  \captionsetup{skip=1pt}
   \caption{Qualitative visualizations of two methods on real-world transparent liquid data.}
   \label{fig:real_result}
\end{figure}


\subsection{Additional Cases of Qualitative Comparison with Reconstruction Baselines}
\label{sec:resultSup}

We selected 10 objects from the test set and randomly chose a single-view scene image for each object. Using our pipeline, we performed reconstruction inference and rendered the reconstructed mesh files from six orthographic views. The reconstruction results are provided as a video in the Supplementary Material, named \textbf{3D Reconstruction Results.mp4}. This video further demonstrates the high-fidelity reconstruction capabilities of the Phys-Liquid pipeline for various types of liquid deformations caused by physical manipulations.

In experiment "Comparisons with Reconstruction Baselines", we compared our pipeline with two other reconstruction models: InstantMesh~\cite{xu2024instantmesh} and TripoSR~\cite{tochilkin2024triposr}. In Figure~\ref{fig:fig13}, we supplement two cases, each from a different object in the test set, using a randomly selected input image. TripoSR~\cite{tochilkin2024triposr} performs worse than InstantMesh~\cite{xu2024instantmesh}, almost failing to capture the 3D geometric features. This comparison exhibits high consistency with the simulation results across six views.

\subsection{Segmentation Model Evaluation}
\label{sec:8_2}
To assess the performance of the segmentation model, we compared the liquid masks generated by YOLO-world~\cite{cheng2024yolo} and SAM2~\cite{ravi2024sam} with the physics-informed simulation masks from the Phys-Liquid dataset using the IoU metric. The segmentation model achieved an average IoU of 0.8694 in Table \ref{tab:tab2}, demonstrating effective segmentation of transparent liquids in complex laboratory environments, establishing a foundation for next three stages.

\begin{table}[ht]
  \centering
  
  \footnotesize 
  \setlength{\tabcolsep}{4pt} 
  \resizebox{\columnwidth}{!}{ 
  \begin{tabular}{@{}lclclc@{}}
    \toprule
    Object & IoU & Object & IoU & Object & IoU \\
    \midrule
    cube-bottle-L & 0.8770 & tube-S & 0.8481 & cylinder-G & 0.6899 \\
    cube-bottle-S & 0.8721 & rect-bottle-L & 0.8226 & cone-tube & 0.7693 \\
    cone-flask-L & 0.9027 & rect-bottle-S & 0.8997 & sphere-flask & 0.8805 \\
    cone-flask-M & 0.8738 & cylinder-L & 0.8802 & cube-flask & 0.9024 \\
    cone-flask-S & 0.8652 & cylinder-S & 0.8873 & cone-bottle & 0.8613 \\
    tube-L & 0.8661 & cylinder flask & 0.8994 & rect-flask & 0.8788 \\
    tube-M & 0.8404 & cylinder-tube & 0.8594 \\
    \midrule
    \multicolumn{6}{c}{Overall Average IoU: 0.8694} \\
    \bottomrule
  \end{tabular}
  }
  \caption{IoU results for 20 objects in the Phys-Liquid dataset.}
  \label{tab:tab2}
\end{table}

\subsection{Mesh Scaling Model Performance Evaluation}
\label{sec:8_3}
We introduce the implementation of the scaling model which utilizes Adam as the optimizer, with learning rates of \(1 \times 10^{-5}\) for the last 6 unfrozen ViT layers and \(1 \times 10^{-4}\) for the view transformer and fully connected layers, using a batch size of 256 with gradient accumulation. We evaluated the mesh scaling model's accuracy in predicting scaling factors $s$ for adjusting reconstructed meshes to real-world dimensions. By comparing the predicted scaling factors $s$ with the physics-informed simulated factors, we calculated the MAPE of $s$. The ViT model achieved a MAPE of 1.75\% in training data and 6.26\% in testing data, indicating high precision in scaling factor prediction.

We present the training and validation processes of two models in our method: the diffusion model for generating multi-view masks and the mesh scaling model. Figure~\ref{fig:fig14} shows the loss curve during the fine-tuning process of the diffusion model. Figure~\ref{fig:fig15} displays the training and validation loss curves of the mesh scaling model using three different loss functions: L1 loss, L2 loss, and Huber loss. Additionally, Figure~\ref{fig:fig16} presents the MAPE of the scaling factor \( s \), illustrating the trend of MAPE over 500 epochs during the training and validation processes.

\subsection{Multi-View Input on Reconstruction Accuracy}
\label{sec:8_4}
As illustrated in pipeline framework, our pipeline supports input images from single view and multi-views for 3D reconstruction of liquids. We investigated the impact of using single-view versus multi-view inputs on the accuracy of the 3D reconstruction. We reconstructed 3D meshes using both single-view and multi-view inputs from a subset of the Phys-Liquid dataset and calculated the RMSE of the mesh dimensions compared to the simulated meshes. The multi-view inputs achieved RMSE levels comparable to those obtained with single-view inputs shown in Table \ref{tab:tab4}, demonstrating that extra viewpoints are not necessarily improving the accuracy.

\begin{table}[ht]
  \centering
  \scriptsize 
  \resizebox{\columnwidth}{!}{ 
    \begin{tabular}{@{}lcccc@{}}
      \toprule
      Combination & RMSE & Chamfer Distance & Volume IoU & F-Score (\%) \\
      \midrule
      Single-view  & 0.0120 & 0.0075 & 0.5086 & 72.10 \\
      Two-view     & 0.0138 & 0.0080 & 0.5310 & 68.57 \\
      Three-view   & 0.0129 & 0.0070 & 0.4933 & 77.14 \\
      \bottomrule
    \end{tabular}
  }
  \caption{Comparative results of reconstructed real-world meshes using input images from single-view, two-view (front and right), and three-view (front, right, and top) configurations.}
  \label{tab:tab4}
\end{table}

\subsection{Visualizations of the reconstructed results tested on the real-world DTLD dataset}

We selected one random case (cylindrical transparent container: G-Rex) from the real-world DTLD~\cite{wang2024towards} dataset and present the qualitative visualizations of rendered multi-view images of reconstruction results by our Phys-Liquid pipeline and Eppel et al.~\cite{eppel2022predicting} method. The results illustrated in Figure~\ref{fig:real_result} include original RGB image, multi-view renderings of the reconstructed meshes by two methods. 

\subsection{Ablation Study on Loss Function}
\label{sec:8_5}
An ablation study was conducted to explore the effect of different loss functions on the performance of the mesh scaling model. We trained the model using L1 loss, L2 loss, and Huber loss~\cite{huber1992robust} and compared their MAPE of scaling factors $s$ recorded in Table \ref{tab:tab5}. The model trained with Huber loss~\cite{huber1992robust} exhibited the highest MAPE during both the training and validation processes. The model trained with L2 loss attained lowest MAPE on the training and validation set with a more stable convergence.

\begin{table}[ht]
  \centering
  \resizebox{\columnwidth}{!}{ 
  \begin{tabular}{lcc}
    \toprule
    Loss & Training MAPE & Validation MAPE \\
    \midrule
    L1 Loss     & 0.0176 & 0.0863 \\
    L2 Loss     & 0.0102 & 0.0638 \\
    Huber Loss  & 0.0197 & 0.0919 \\
    \bottomrule
  \end{tabular}
  }
  \caption{The MAPE results using L1, L2, and Huber Loss~\cite{huber1992robust}.}
  \label{tab:tab5}
\end{table}


\end{document}